%% file: main.tex
\documentclass[10pt,twocolumn,letterpaper]{article}

\usepackage[pagenumbers]{iccv} 


\definecolor{iccvblue}{rgb}{0.21,0.49,0.74}
\usepackage[pagebackref,breaklinks,colorlinks,allcolors=iccvblue]{hyperref}
\usepackage{xcolor}
\usepackage{lipsum}
\usepackage{booktabs}
\usepackage{graphicx}
\usepackage{colortbl}
\usepackage{makecell}
\usepackage{arydshln} 
\definecolor{lightgray}{rgb}{0.7,0.7,0.7} 
\usepackage{multirow}

\usepackage{xr}
\def\paperID{5678} 
\def\confName{ICCV}
\def\confYear{2025}

\title{VLM4D: Towards Spatiotemporal Awareness in Vision Language Models
}

\author{Shijie Zhou$^1$\thanks{Equal contribution.}\quad Alexander Vilesov$^1$\footnotemark[1]\quad Xuehai He$^{2,3}$\footnotemark[1]\quad Ziyu Wan$^2$\quad Shuwang Zhang$^1$\; \\[0.2em] Aditya Nagachandra$^1$ \; Di Chang$^4$ \; Dongdong Chen$^2$ \; Xin Eric Wang$^3$ \; Achuta Kadambi$^1$\\[0.4em]
\normalsize{$^1$UCLA \quad
$^2$Microsoft \quad
$^3$UCSC \quad
$^4$USC}\\[0.3em]
\url{https://vlm4d.github.io/}
}

\begin{document}
\twocolumn[{%
\renewcommand\twocolumn[1][]{#1}%
\maketitle
    \captionsetup{type=figure}
    \vspace{-9mm}
    \includegraphics[width=\textwidth]{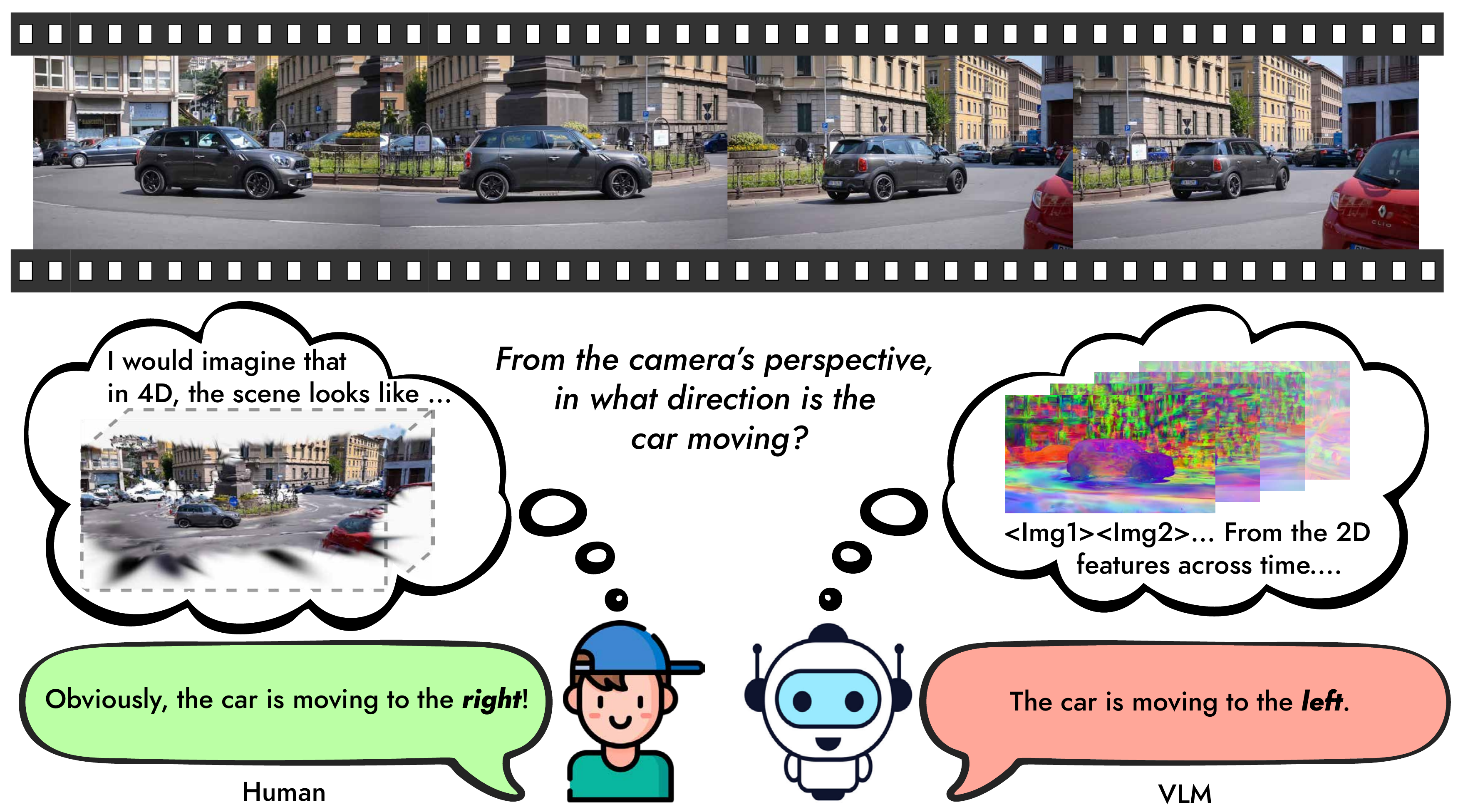} 
    \vspace{-5mm}
    \hfill\caption{\textbf{Spatiotemporal (4D) Awareness}. Humans intuitively reason in 4D (3D space + time), effortlessly reconstructing the dynamic spatial trajectory of moving objects from any perspective. In contrast, current Vision Language Models (VLMs) typically rely on aggregating 2D visual features across time, leading to incorrect predictions when motion understanding and interpretation requires deeper spatiotemporal reasoning. In this example, humans correctly perceive the car moving to the right, while the VLM (GPT-4o) inaccurately predicts leftward movement, suggesting VLMs struggle to perform spatiotemporal reasoning.}
    \label{fig:teaser}
    \hfill \vspace{0mm}
}]
{
\renewcommand{\thefootnote}{\fnsymbol{footnote}}
\footnotetext[1]{Equal contribution.}
}

\input{sec/0_abstract}  
\input{sec/1_intro}

\input{sec/2_related}
\input{sec/3_benchmark}
\input{sec/4_evaluation}

\input{sec/5_solutions}

\input{sec/6_conclusion}

\clearpage
{
    \small
    \bibliographystyle{ieeenat_fullname}
    \bibliography{main}
}
\input{sec/X_suppl}

\end{document}

%% file: sec/0_abstract.tex
\begin{abstract}
Vision language models (VLMs) have shown remarkable capabilities in integrating linguistic and visual reasoning but remain fundamentally limited in understanding dynamic spatiotemporal interactions. Humans effortlessly track and reason about object movements, rotations, and perspective shifts-abilities essential for robust dynamic real-world understanding yet notably lacking in current VLMs. In this paper, we introduce \texttt{VLM4D}, the first benchmark specifically designed to evaluate the spatiotemporal reasoning capabilities of VLMs. Our benchmark comprises diverse real-world and synthetic videos accompanied by carefully curated question-answer pairs emphasizing translational and rotational motions, perspective awareness, and motion continuity. Through comprehensive evaluations of state-of-the-art open and closed-source VLMs, we identify significant performance gaps compared to human baselines, highlighting fundamental deficiencies in existing models. Extensive analysis reveals that VLMs struggle particularly with integrating multiple visual cues and maintaining temporal coherence. We further explore promising directions, such as leveraging 4D feature field reconstruction and targeted spatiotemporal supervised fine-tuning, demonstrating their effectiveness in enhancing spatiotemporal comprehension. Our work aims to encourage deeper exploration into improving VLMs' spatial and temporal grounding, paving the way towards more capable and reliable visual intelligence for dynamic environments.
\end{abstract}

%% file: sec/1_intro.tex
\section{Introduction}
\label{sec:intro}
Humans possess an innate ability to perceive, track, and interpret motion as well as spatial and temporal changes, enabling rich interpretations of complex dynamic events from both egocentric and allocentric perspectives \cite{DeFreitas2016Tracking, Marr1981Directional, Burgess2006Spatial}. When observing an object move, we can inherently process any changes such as lateral shifts, rotational directions, and periodic or repeated actions unfolding along a specific trajectory~\cite{Burgess2006Spatial}. These sophisticated perceptual abilities are a product of our spatiotemporal cognition~\cite{Freyd1984Representational}, and form an essential foundation that allows us to comprehend and reason about physical phenomena, object interactions, and causal relationships within our environment~\cite{Leslie1984Spatiotemporal, Johansson1973Visual}.

Vision language models (VLMs), which also have the potential to perceive motion and spatiotemporal changes in videos, represent a prominent class of methods aimed at emulating or surpassing human capabilities in integrated visual and linguistic reasoning~\cite{LeCun1989Backpropagation, Dosovitskiy2021An}.
While previous work on VLMs has primarily focused on static visual understanding—through large-scale training on paired language and image data~\cite{Radford2021Learning}—or explored video understanding tasks such as captioning~\cite{Lu2019ViLBERTPT} and scene understanding~\cite{Chen_2022_CVPR}, we find that their strong performance in these tasks does not naturally translate into robust spatiotemporal reasoning. This limitation is particularly striking given that state-of-the-art VLMs are typically trained on datasets comprising hundreds of billions of tokens~\cite{liu2023world}. In contrast, human infants naturally develop robust spatiotemporal cognition within the first few months of life~\cite{Spelke2007Core}. 

Another key challenge that limits VLM performance on spatiotemporal tasks is the need to implicitly or explicitly reconstruct a four-dimensional (4D) representation—3D space + time—of dynamic scenes, and subsequently reason over this reconstruction~\cite{wang2024compositional}. As illustrated in~\cref{fig:teaser}, the car is advancing forwards and turning to the left in its own frame of reference. However, from the camera's perspective, its motion appears as a combination of heading to the right and receding into the distance despite the car being in the center of the frame due to camera rotation. Human observers can seamlessly disentangle these complex dynamics, accurately interpreting trajectories by synthesizing diverse visual cues including camera rotation compensation, stationary scene landmarks, prior knowledge of 3D and 4D environmental structures, and perspective projections~\cite{Marr1981Directional,Burgess2006Spatial,Leslie1984Spatiotemporal,Freyd1984Representational}. 
The inability of current VLMs to integrate spatial, temporal, and semantic cues in a human-like manner stems from their fundamentally different encoding paradigms for visual and language information. This discrepancy highlights a significant gap between human and machine spatiotemporal understanding, suggesting that future VLMs may benefit from insights in cognitive science and neuroscience to develop more advanced mechanisms for perceiving, integrating, reconstructing, and reasoning over dynamic scenes.

\begin{figure}[t]
    \centering
    \includegraphics[width=0.45\textwidth]{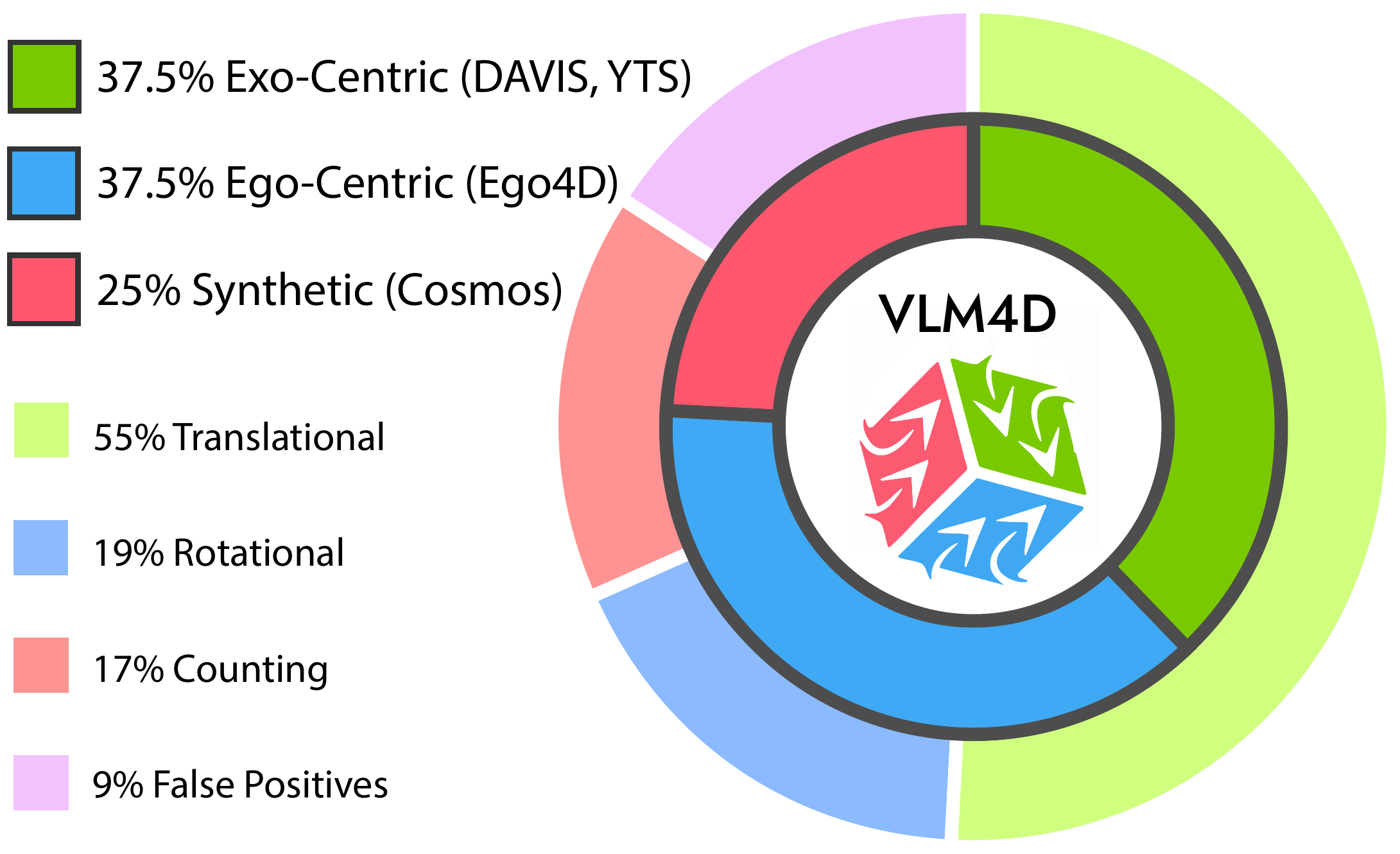}
    \caption{\textbf{Distribution of Dataset Sources and Annotations.} Overview of the dataset composition, illustrating the proportions of real third-person (exo-centric) videos (DAVIS~\cite{davis2017}, YouTube-VOS~\cite{xu2018youtube}), real first-person (ego-centric) videos (Ego4D~\cite{grauman2022ego4d}), and synthetic videos (Cosmos~\cite{agarwal2025cosmos}). The real video data is further categorized by annotation types, including translational, rotational, action, counting, and false positive queries (targeting nonexistent events to assess critical reasoning).}
    \label{fig:data_piechart}
    \vspace{-0.3cm}
\end{figure}

\begin{figure*}[ht]
    \centering
    \includegraphics[width=1\textwidth]{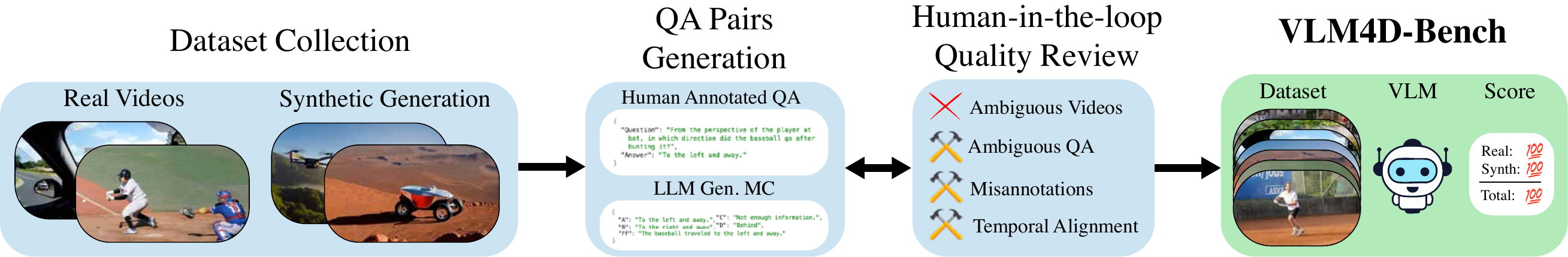}
    \caption{\textbf{Dataset Generation and Annotation Pipeline.} Our dataset was constructed by collecting real videos and generating synthetic data, followed by human-in-the-loop quality reviews to address ambiguous videos and annotations. After temporal alignment and quality assurance, human-annotated questions and ground-truth answers were created, complemented by multiple-choice (MC) answers generated by large language models (LLMs). The final dataset includes real and synthetic video data with comprehensive VLM scoring metrics.}
    \label{fig:dataset_pipeline}
    \vspace{-0.4cm}
\end{figure*}

In order to effectively characterize and challenge the existing spatiotemporal reasoning abilities of VLMs, we introduce \texttt{VLM4D}, a rigorous benchmark specifically designed to probe the spatiotemporal grounding capabilities of current vision language models. Through this contribution, we aim to catalyze research that addresses the critical gap in spatiotemporal understanding and reasoning within VLMs and provide a foundational analysis highlighting key deficiencies in existing models. 

We summarize our main~\textbf{contributions} as follows:
\begin{enumerate}
\item We propose the first benchmark~\texttt{VLM4D} explicitly designed to rigorously evaluate the spatiotemporal (4D) reasoning capabilities of Vision Language Models (VLMs).
\item We introduce a novel, meticulously curated dataset consisting of diverse real-world and synthetic video sequences paired with carefully crafted spatiotemporal question-answer (QA) annotations.
\item We analyze the critical limitations of contemporary VLMs in spatiotemporal reasoning and conduct experiments exploring potential solutions, highlighting fundamental challenges and outlining clear directions for impactful future research.
\end{enumerate}

%% file: sec/2_related.tex
\section{Related Work}
\label{sec:related_work}

\paragraph{Spatiotemporal Understanding in Vision Language Models}%
Early methods for video understanding before the advent of large vision language models (VLMs) leveraged trajectory-based representations and separate modeling of spatial and motion cues for tasks like action and motion recognition~\cite{Laptev2005, Wang2011, Simonyan2014, chari2023learning}. Recently, VLMs have evolved rapidly by fully leveraging the significant achievements of Large Language Models (LLMs)~\cite{brown2020language, devlin2019bert, wei2021finetuned, bai2023qwen, touvron2023llama, radford2018improving} and large-scale visual instruction tuning datasets~\cite{liu2023visual, zhu2023minigpt, dai2023instructblip}. While VLMs~\cite{gong2023multimodal, liu2023visual, zhu2023minigpt, abouelenin2025phi4minitechnicalreportcompact, hurst2024gpt, li2024llava, team2024gemini, wang2024qwen2} exhibit transformative potential for applications such as embodied AI~\cite{suglia2024alanavlm, driess2023palm, kim2024openvla}, robotics~\cite{wang2024vlm, patel2025real}, scene generation~\cite{ocal2024sceneteller, zhou2024dreamscene360, ling2025scenethesis}, and world modeling~\cite{liu2024world, zhang2024combo}, most existing methods remain constrained to static images, focusing narrowly on spatial understanding~\cite{cheng2024spatialrgpt, chen2024spatialvlm, ranasinghe2024learning, fan2025vlm} while overlooking the dynamic temporal dimension inherent in real-world interactions. To bridge this gap, emerging research~\cite{li2023videochat, zhang2023video, cheng2024videollama, maaz2023video, zhang2025videollama} has begun exploring video modality integration, aiming to equip VLMs with spatial-temporal awareness critical for tasks like video comprehension, where both contextual details and motion dynamics are essential. For example, VideoLLM-MoD~\cite{wu2024videollm} proposes to address the efficiency issue when processing long-term video by mixture-of-depths. ~\cite{yuan2024videorefer} introduces VideoRefer to enhance the finer-level (like object-level) spatial-temporal video understanding of VLMs. Grounded-VideoLLM~\cite{wang2024grounded} also targets for fine-grained video understanding through incorporating an additional temporal stream. In this work, we aim to rigorously evaluate the 4D spatial-temporal reasoning capabilities of state-of-the-art VLMs, probing how and to what extent these models internalize spatial intelligence and temporal dependencies.

\paragraph{VLM Benchmarks}%
Following the development trends of VLMs, benchmarking VLMs shares the similar trajectory by first evaluating vision QA on static images~\cite{li2024seed, liu2024mmbench, he2024mmworld,yue2024mmmu}, to align with models’ early focus on 2D understanding. As VLMs evolved to tackle dynamic scenarios, benchmarks expanded to evaluate general-purpose video comprehension tasks that probe temporal coherence and event understanding~\cite{ning2023video, khattak2024good, li2024videoeval, fu2024video, li2024mvbench}. Notably, MMVU~\cite{zhao2025mmvu} further proposes a knowledge-intensive benchmark to assess the expert-level reasoning ability of current video-based large models. However, while these works assess perception and semantic understanding, they largely overlook the explicit evaluation of spatial-temporal awareness, a core capability for real-world applications requiring 4D (3D space + time) reasoning. Recent efforts like~\cite{yang2024thinking} pioneer benchmarks for 3D visual-spatial intelligence but restrict evaluation to static 3D scene, neglecting the interplay of object motion and temporal dynamics intrinsic to videos. In this work, we introduce \texttt{VLM4D}, the first benchmark designed to holistically evaluate the 4D intelligence in VLMs, unifying spatial understanding, temporal continuity, and motion reasoning. By curating tasks that demand precise analysis of dynamic interactions (e.g., direction prediction, perspective anticipation, and motion reasoning), \texttt{VLM4D} exposes critical gaps in current models' ability to internalize spatiotemporal relationships. Our work not only advances the granularity of VLM evaluation but also shares insights and potential solutions to improve the model performance.

\begin{figure*}[t]
    \centering
    \includegraphics[width=0.8\textwidth]{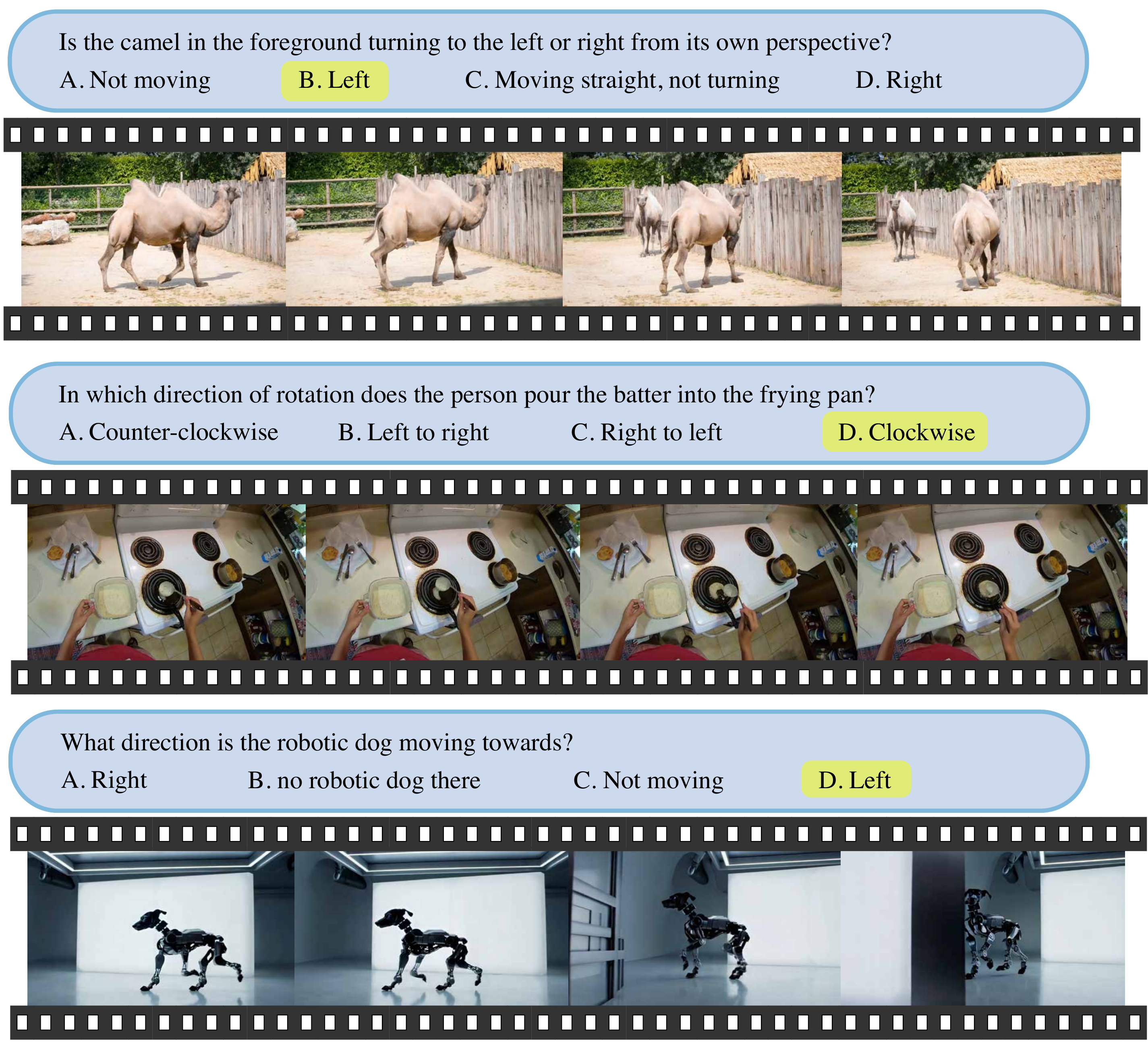}
    \caption{\textbf{Qualitative Examples of Dataset Annotations.} (Top) A third-person (exo-centric) video with translational annotations (``camel turning left from its perspective"). (Middle) A first-person (ego-centric) video with a rotational question (``clockwise rotation of ladle"). (Bottom) A synthetic scene with motion recognition ``robotic dog moving left"). }
    \label{fig:qualitative_examples}
    \vspace{-0.5cm}
\end{figure*}

%% file: sec/3_benchmark.tex
\section{The \texttt{VLM4D} Benchmark}
We introduce \texttt{VLM4D}, the first benchmark specifically designed to test the spatiotemporal reasoning ability of VLMs. \texttt{VLM4D} consists of 1000 videos paired with over 1800 question-answer pairs, each carefully designed to assess both spatial and temporal understanding jointly. The majority of these videos are sourced from datasets with rich spatiotemporal characteristics, thus ensuring a diverse range of motion-related scenarios. We also augment the dataset with synthetic videos generated by a world-foundation model, Cosmos~\cite{agarwal2025cosmos}, that has been modified using techniques introduced in~\cite{he2024mojito} to obtain more accurate correspondence between motion-oriented prompts and the resulting generated video. \cref{fig:data_piechart} illustrates the composition of our dataset.

\subsection{Benchmark Construction}
Unlike prior work that often relies heavily on LLMs and VLMs to generate first iterations of benchmarks and datasets~\cite{chen2024sharegpt4video} followed by human quality control, we found that existing VLMs and automated methods showed significant limitations in terms of reliability and quality. This shortcoming necessitated direct human annotations that were then followed by augmentation by LLMs to ensure a high-quality benchmark. An overview of the benchmark curation pipeline is shown in \cref{fig:dataset_pipeline}. 

\paragraph{Real Video Data Collection} Real-world videos were sourced from datasets with rich spatiotemporal characteristics that ensured diverse motion and perspective variations. For egocentric data, we relied mainly on the Ego4D dataset~\cite{grauman2022ego4d}, while most exocentric data points were collected from the DAVIS~\cite{davis2017} and YouTube-VOS~\cite{xu2018youtube} datasets. To minimize confounders and to focus attention of VLM abilities to only spatiotemporal reasoning, we preprocessed the videos by temporally segmenting and centering them around the most relevant action, thus resulting in videos with an average duration of $3$-$8$ seconds. This ensures that the key event described in the question is clear and reduces ambiguities or confounders that would reduce VLM accuracy. 

\paragraph{Synthetic Video Data Generation}  
The rapid advancement of video generation techniques has led to their widespread application in diverse domains like robotics and immersive entertainment~\cite{agarwal2025cosmos, videoworldsimulators2024, wan2025wan, li2025kdgen, wu2025cat4d}, making 4D reasoning on synthetic content essential for VLMs and underscoring the need to include such videos in our benchmark. For synthetic video generation, we use Cosmos~\cite{agarwal2025cosmos} as our video generation backbone. To ensure that the generated videos align with the intended object moving directions, we incorporate input bounding boxes as additional spatial guidance. Specifically, we follow the approach introduced in~\cite{he2024mojito} modifying the diffusion forward steps to enforce object localization constraints at each timestep, ensuring consistency between the generated object direction and the user-specified trajectory. The average duration of generated synthetic videos is $5$ seconds. To maintain high-quality outputs, we perform a manual verification step after generation, filtering out low-quality videos and retaining only those that accurately match the specified directions. Once a video is generated, we use an LLM (GPT-4o) to create two types of evaluation questions: Directional questions, derived from the textual prompt used to generate the video; and False Positive (FP) questions (constituting 10\%, matching the ratio in the real dataset), which query non-existent objects in the scene. Both question types follow the format: ``What direction is the $\langle$Object Name$\rangle$ moving?", where the model must select one of four possible answers: ``left", ``right", ``not moving", or ``no $\langle$Object Name$\rangle$ there". A final manual review is conducted to filter out or revise ambiguous questions, ensuring the quality of the questions and ground-truth answers.

\paragraph{QA Generation and Quality Control} Question-answer pairs are primarily constructed through human annotations. The question answer pairs are then supplemented with alternative answers by an LLM (GPT-4o) for multiple choice (MC) questions. To ensure high-quality annotations, we applied a three-round, multi-person cross-checked verification process, during which ambiguous videos were filtered out and vague, misleading, or incorrect QA pairs were refined to improve spatial and temporal alignment between language and visual content. \cref{fig:qualitative_examples} showcases some qualitative examples of annotations for different types of videos.

\paragraph{Assessing Human Performance}
To establish a human performance baseline on our benchmark, we conducted an evaluation in which participants independently answered 100 randomly sampled questions from the dataset. The accuracy of human responses was then aggregated to approximate the performance of human spatiotemporal reasoning on the dataset. 

\subsection{Categorizing Spatiotemporal Performance}
\begin{figure}
    \centering
    \includegraphics[width=1\linewidth]{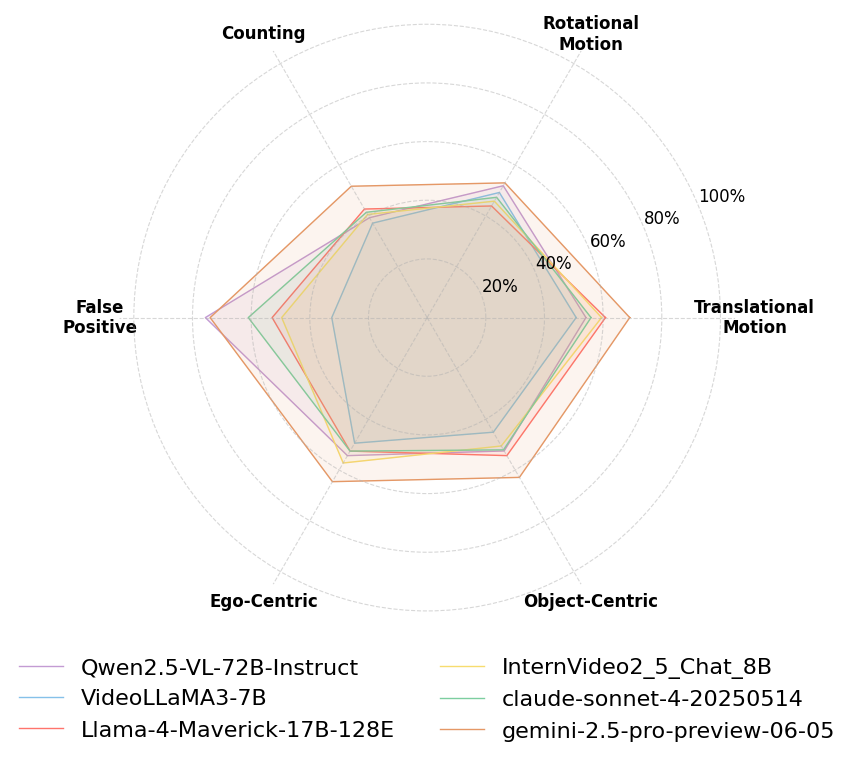}
    \caption{\textbf{Comparison of accuracy across types of spatiotemporal questions.} Model accuracy is shown only for the six top-performing VLMs.}
    \label{fig:radar_plot}
    \vspace{-0.3cm}
\end{figure}

To systematically evaluate spatiotemporal reasoning capabilities, we first categorize videos into two primary groups: egocentric (first-person) videos and exocentric (third-person) videos. Egocentric videos are sourced from the Ego4D~\cite{grauman2022ego4d} dataset, where scenes are captured from a head-mounted camera, thus offering dynamic video data that is inherently coupled with the individual's actions. Exocentric videos encompass a diverse range of recorded scenes, from sports footage to everyday scenes. Beyond this categorization, we also evaluate spatiotemporal performance across four dimensions: translational movement (TM), rotational movement (RM), spatiotemporal counting (STM), and false positives (FP), with their proportions shown in \cref{fig:data_piechart}. Translational movement assesses a model's ability to track linear motion within scenes, while rotational movement assesses the understanding of changes in orientation and perspective shifts over time. Spatiotemporal counting extends these core motion-based tasks by requiring a more complex reasoning strategy to determine the number of objects performing a translation or rotational movement. Lastly, the false positives category evaluates the model’s critical thinking~\cite{fan2025missing} in determining whether an object or event actually occurred within the spatiotemporal context. By structuring the benchmark along these axes, we aim for a comprehensive framework for assessing spatiotemporal reasoning (\cref{fig:radar_plot}).

%% file: sec/4_evaluation.tex
\begin{table*}[h]
    \centering
    \resizebox{1.0\linewidth}{!}{
    \renewcommand{\arraystretch}{1.2}
    \arrayrulecolor{black} 
    \begin{tabular}{l l l c c c c c c c}
    \toprule
    \textbf{Organization} & \textbf{Model} & \textbf{Release} & \multicolumn{3}{c}{\textbf{Real}} & \multicolumn{3}{c}{\textbf{Synthetic}} & \textbf{Overall} \\
    \cmidrule(lr){4-6}
    \cmidrule(lr){7-9}
    & & & \textbf{Ego-centric} & \textbf{Exo-centric} & \textbf{Average} & \textbf{Directional} & \textbf{FP} & \textbf{Average} \\
    \midrule
         User Study & Human Performance &  & 99.6 & 99.7 & 99.7 & 95.8 & 100 & 96.2  & 98.8     \\
         Random & Random Selection  &  & 24.4 & 23.2 & 23.6 & 25.5 & 24.7 & 25.4 & 24.1    \\
        \midrule
        \rowcolor[HTML]{e1f0f5}\multicolumn{10}{l}{\textbf{Latest Proprietary VLMs}} \\
        \midrule
        OpenAI & GPT-4o & 2024-11 & \cellcolor[HTML]{FFE5CC}{55.5} & \cellcolor[HTML]{FFCC99}{62.2} & \cellcolor[HTML]{FFCC99}{60.0} & \cellcolor[HTML]{FFE5CC}{49.5} & 53.3 & 49.9 & \cellcolor[HTML]{FFCC99}{57.5} \\
        \arrayrulecolor{lightgray} \hdashline
        Google & Gemini-2.5-Pro & 2025-6 & \cellcolor[HTML]{FFB366}{64.6} & \cellcolor[HTML]{FFB366}{62.9} & \cellcolor[HTML]{FFB366}{63.5} & \cellcolor[HTML]{FFB366}{54.8} & \cellcolor[HTML]{FFCC99}{80.0} & \cellcolor[HTML]{FFB366}{57.3} & \cellcolor[HTML]{FFB366}{62.0} \\
        \hdashline
        Anthropic & Claude-Sonnet-4 & 2025-5 & 52.6 & 52.1 & 52.2 & 44.0 & \cellcolor[HTML]{FFB366}{86.7} & 48.3 & 51.3 \\
        \hdashline
        xAI & Grok-2-Vision & 2024-12 & 48.8 & 49.7 & 49.4 & 49.3 & 66.7 & 51.0 & 49.8 \\
        \arrayrulecolor{black} \midrule
        \rowcolor[HTML]{e1f0f5}\multicolumn{10}{l}{\textbf{Open-source Image VLMs}} \\
        \midrule
        Meta 
        & Llama-4-Maverick-17B & 2025-4 & 52.6 & 54.3 & \cellcolor[HTML]{FFE5CC}{53.8} & \cellcolor[HTML]{FFCC99}{53.3} & 51.1 & \cellcolor[HTML]{FFE5CC}{53.0} & 53.6 \\
        & Llama-4-Scout-17B & 2025-4 & 48.6 & \cellcolor[HTML]{FFE5CC}{56.2} & 53.7 & \cellcolor[HTML]{FFCC99}{53.3} & \cellcolor[HTML]{FFE5CC}{75.6} & \cellcolor[HTML]{FFCC99}{55.5} & \cellcolor[HTML]{FFE5CC}{54.1} \\
        \arrayrulecolor{lightgray} \hdashline
        Microsoft & Phi-4-Multimodal & 2025-3 & 41.0 & 35.4 & 37.2 & 37.5 & 11.1 & 34.8 & 36.6 \\
        & Phi-3.5-Vision & 2024-7 & 33.4 & 38.8 & 37.1 & 23.3 & 37.8 & 24.7 & 34.0 \\
        \arrayrulecolor{lightgray} \hdashline
        DeepSeek &  DeepSeek-VL2 & 2024-12 & 33.6 & 32.9 & 33.1 & 31.8 & 46.7 & 33.3 & 33.2 \\
        \arrayrulecolor{lightgray} \hdashline
        Shanghai AI Lab & InternVL2.5-38B & 2024-11 & 46.6 & 50.1 & 48.9 & 43.3 & 57.8 & 44.7 & 47.9 \\
        & InternVL2.5-8B & 2024-11 & 39.0 & 44.0 & 42.4 & 40.8 & 42.2 & 40.9 & 42.0 \\
        \arrayrulecolor{lightgray} \hdashline
        Mistral AI & Pixtral-12B & 2024-9 & 32.3 & 25.8 & 27.9 & 24.3 & 22.2 & 24.0 & 27.0 \\
        \arrayrulecolor{lightgray} \hdashline
        Rhymes & Aria & 2024-11 & 47.2 & 44.0 & 45.1 & 38.5 & 71.1 & 41.8 & 44.3 \\
        \arrayrulecolor{black} \midrule
        \rowcolor[HTML]{e1f0f5} \multicolumn{10}{l}{\textbf{Open-source Video VLMs}} \\
        \midrule
        Alibaba & Qwen2.5-VL-7B & 2025-1 & 42.3 & 43.7 & 43.3 & 43.5 & 64.4 & 45.6 & 43.8 \\
        & Qwen2.5-VL-72B & 2025-1 & 54.3 & 52.5 & 53.1 & \cellcolor[HTML]{FFE5CC}{49.5} & \cellcolor[HTML]{FFCC99}{80.0} & 52.6 & 53.0 \\
        & Qwen2-VL-7B & 2024-8 & 36.1 & 34.7 & 35.2 & 40.5 & 35.6 & 40.0 & 36.3 \\
        & Qwen2-VL-72B & 2024-9 & 48.1 & 43.0 & 44.6 & 40.8 & 73.3 & 44.0 & 44.5 \\
        \hdashline
        DAMO & VideoLLama3-2B & 2025-1 & 53.2 & 42.5 & 46.0 & 34.3 & 55.6 & 36.4 & 43.7 \\
        & VideoLLama3-7B & 2025-1 & 49.4 & 45.1 & 46.5 & 42.8 & 53.3 & 43.8 & 45.9 \\
        \hdashline
        Shanghai AI Lab & InternVideo2.5-8B & 2025-1 & \cellcolor[HTML]{FFCC99}{57.2} & 50.5 & 52.7 & 44.3 & 46.7 & 44.5 & 50.7 \\
        & InternVideo2-8B & 2024-8 & 35.6 & 39.3 & 38.1 & 43.0 & 0.0 & 38.7 & 38.2\\
        \hdashline
        LLaVA & LLaVA-One-Vision-7B & 2024-9 & 36.8 & 35.6 & 36.0 & 37.8 & 35.6 & 37.5 & 36.3  \\
        & LLaVA-NeXT-Video-34B & 2024-6 & 29.6 & 31.6 & 30.9 & 24.5 & 55.6 & 27.6 & 30.1\\
        \arrayrulecolor{black} \bottomrule
    \end{tabular}
    }
    \caption{\textbf{Evaluation on \texttt{VLM4D} Benchmark} across various proprietary and open-source VLMs. Top three performers in each category are highlighted from \colorbox[HTML]{FFB366}{dark} (highest) to \colorbox[HTML]{FFE5CC}{light} (third highest). Human and random selection baselines are included for reference.} 
\label{tab:result}
\end{table*}  

\section{Evaluation of \texttt{VLM4D} Benchmark}
\subsection{Evaluation Setup}
\paragraph{Benchmark Models}
We evaluate 23 most recently released VLMs thus covering a wide range of model sizes, architectures, and training methodologies. For closed-source VLMs, we evaluate GPT-4o~\cite{gpt4o}, Gemini 2.5 Pro~\cite{gemini2_5}, Claude Sonnet 4~\cite{claude4}, and Grok-2-Vision~\cite{grok2}. For open-source models, we include Llama 4~\cite{llama4}, DeepSeek-VL~\cite{lu2024deepseek}, Qwen2.5-VL~\cite{yang2024qwen2}, Qwen2-VL~\cite{wang2024qwen2}, InternVL2.5~\cite{chen2024expanding}, Aria~\cite{li2024aria}, InternVideo 2.5~\cite{wang2025internvideo2}, InternVideo2~\cite{wang2024internvideo2}, Phi-4-multimodal~\cite{abdin2024phi4}, Phi-3.5-vision~\cite{abdin2024phi3}, Pixtral~\cite{agrawal2024pixtral}, VideoLLama3~\cite{zhang2025videollama}, Llava-One-Vision~\cite{li2024llava}, Llava-NeXT-Video~\cite{zhang2024llavanextvideo}. When available, we evaluate different sizes for each model, resulting in models ranging from 2 to 72 billion parameters. 

\paragraph{Evaluation Settings}
The evaluations were performed in a zero-shot setting with video or a set of sampled frames of video, followed by the prompt forming the input. For each model, we evaluate on two different inference settings. In the first setting, the model prompted to directly output (DO) the answer immediately without any reasoning, and in the second evaluation setting, the model is directed to create intermediate reasoning steps, chain-of-thought (CoT)~\cite{wei2022chain}, before inferring the final answer. 

\paragraph{Metrics} 
Following prior work~\cite{yang2024thinking} and given the nature of our target task, we adopt multiple-choice questions (MCQs) for evaluation, using accuracy as the primary metric. For the two inference settings described earlier, we employ LLM-as-Judge~\cite{zhao2025mmvu} to assess the outputs of VLMs. We opt for this method instead of string or template matching, as VLMs—especially under chain of thought (CoT) prompting—often generate all possible answer choices during reasoning, with varying frequencies and slight formatting differences. In some cases, the final answer may even contradict the reasoning. To better evaluate whether the model truly understands the video, we prompt two advanced LLMs (GPT-o3 and o4-mini) to grade based on the full CoT reasoning, not just the final answer. We then perform a cross-check between their judgments and manually resolve any disagreements. The evaluation results are reported in~\cref{tab:result}.

\subsection{Benchmark Results}
Our comprehensive evaluation on the \texttt{VLM4D} benchmark, detailed in~\cref{tab:result}, systematically assesses the spatiotemporal reasoning capabilities of modern Vision-Language Models (VLMs). The results highlight a clear hierarchy, with proprietary models demonstrating superior performance over their open-source counterparts. Google's Gemini-2.5-Pro emerges as the top-performing model with an overall accuracy of 62.0\%, followed by OpenAI's GPT-4o at 57.5\%. Within the open-source domain, image-based models like Meta's Llama-4-Scout-17B (54.1\%) and video-supported models like Alibaba's Qwen2.5-VL-72B (53.0\%) demonstrate highly competitive results, even outperforming some proprietary counterparts. Despite these achievements, a significant performance gap persists when compared to human accuracy (98.8\%), underscoring that sophisticated 4D awareness remains a formidable challenge for AI. Notably, performance varies significantly across different categories, such as real versus synthetic data and ego-centric versus exo-centric perspectives, indicating that current models lack generalized spatiotemporal understanding.



%% file: sec/5_solutions.tex
\section{Analysis: Why VLMs Don't Work Well?}

\subsection{Limited Spatiotemporal Cognition}
\begin{figure}
    \centering
    \includegraphics[width=\linewidth]{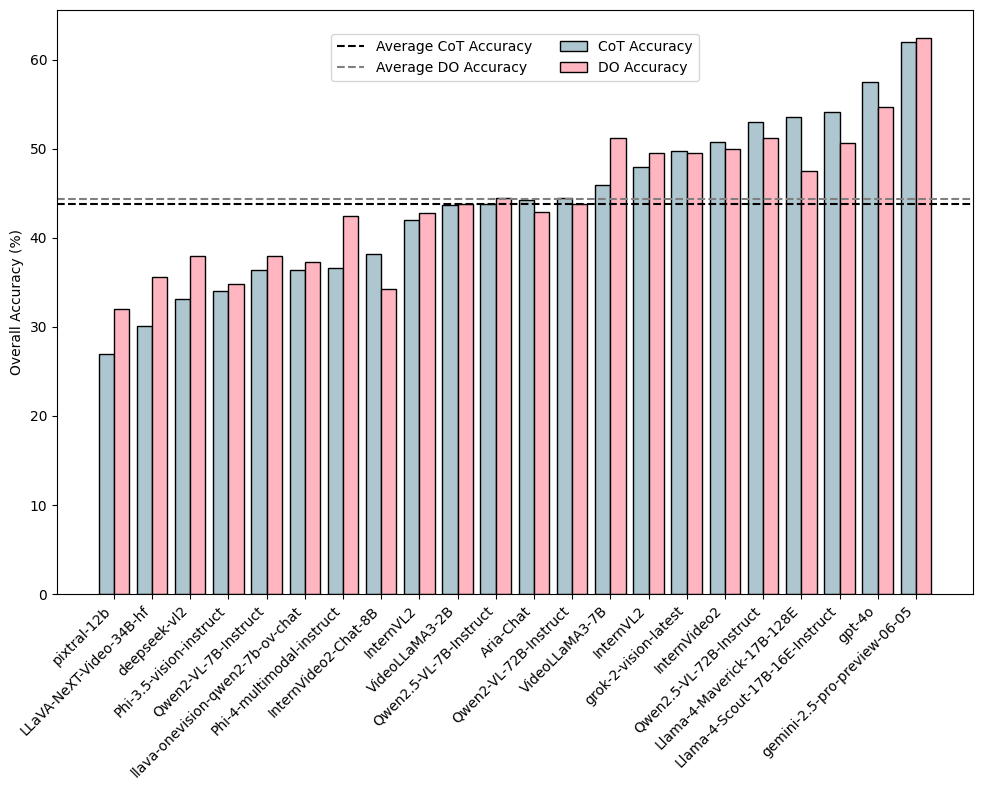}
    \caption{\textbf{Comparison of CoT and DO Accuracy Across Models.} Accuracy comparison between Chain-of-Thought (CoT) and Direct Output (DO) prompting across VLMs.}
    \label{fig:cot_vs_do}
    \vspace{-0.1cm}
\end{figure}

Despite significant advances in VLMs, their abilities to understand and reason about motion, spatial relationships, and temporal coherence remains fundamentally underdeveloped ~\cite{chen2024we, zohar2024apollo}. Chain of thought (CoT)~\cite{wei2022chain} is widely employed as a method to improve accuracy through step-by-step reasoning. We showcase a comparison between CoT and DO in ~\cref{fig:cot_vs_do}. Overall, there is no indication of a large advantage of CoT over all evaluated models. Upon deeper exploration of the CoT reasoning of some models, we observe that the reasoning process was primarily flawed in the following ways: irrelevant information and arriving at conclusions that are inconsistent with the reasoning process. Larger models exhibited strategies that would be similar to how a human processes spatiotemporal information, but the resulting execution falls short of human performance. This demonstrates a disconnect between its visual and linguistic knowledge. We provide examples of this behavior in the supplementary material.

\subsection{Deficiencies in Spatiotemporal Labeling}
\begin{figure}[t]
    \centering
    \includegraphics[width=0.5\textwidth]{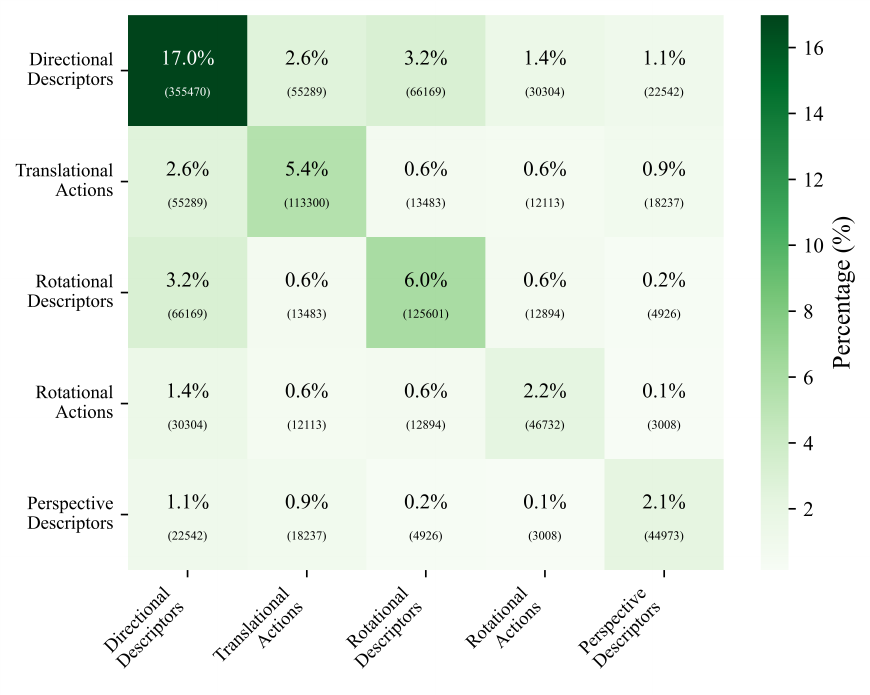}
    \caption{\textbf{Heatmap of Occurances of Spatial-Temporal Terms in popular video SFT datasets.}}
    \label{fig:sft_ds}
\end{figure}

Another avenue of exploration we undertook is to understand the richness of spatiotemporal labels in popular supervised fine-tuning (SFT) VLM datasets. Typically, video captioning occurs at the `scene' level, lacking fine-grained temporal, spatial, and object-level details. We performed an extensive analysis, encompassing over 2 million samples~\cite{chen2024sharegpt4video, cui2025comprehensive, li2023videochat, li2023mvbench, zhang2024llavanextvideo}. We performed this analysis through string-matching of spatiotemporal descriptors related to directionality, translational motion, rotation, and perspective shifts and provide the overall results in ~\cref{fig:sft_ds}. We then performed a manual finegrained evaluation of the ShareGPT4Video dataset \cite{chen2024sharegpt4video} which we found had the highest density of spatiotemporal dataset. We found that from a sample of 100 labels that were detected as spatiotemporal, less than 10\% of them were judged as accurate upon human evaluation. This result underscores the inadequacy of current dense captioning approaches, which frequently generate spatiotemporal descriptors without capturing precise motion dynamics. We provide more detailed analysis and explanations in the supplementary material.

\section{Probing Future Solutions}
To probe promising future solutions for enhancing spatiotemporal video understanding, we propose two approaches that address some of the shortcomings of current state-of-the-art VLMs: fine-tuning a VLM on data-rich in spatiotemporal actions and the other leveraging 4D reconstruction and feature fields jointly with a VLM. SFT refines the model’s abilities by training on datasets that contain temporally and spatially rich actions and interactions. By integrating structured visual representations and targeted fine-tuning, these approaches enhance video-language models’ ability to interpret motion. The second method lifts the feature space of VLMs into a temporally coherent 4D feature field, providing structured scene representations that improve motion and spatial reasoning in the stage of decoding and inference.

\paragraph{Spatial-Temporal SFT}
We evaluate on a subset split of the real dataset by randomly splitting the real-world dataset into a training and testing split (80\% / 20\%) and we try settings using synthetic/real/both for training. We conducted the experiments using Qwen 2VL (7B) and Qwen 2.5VL (7B) through LLama-Factory~\cite{zheng2024llamafactory}, and compared the performance before and after supervised fine-tuning in~\cref{tab:sft}. The results demonstrated an improvement in accuracy in spatiotemporal reasoning, suggesting that performance gains can be obtained through targeted training. However, the addition of synthetic data does not necessarily increase performance over using real data alone, suggesting the importance of synthetic data quality.

\paragraph{4D Feature Fields Reconstruction} Recent advances in 3D/4D feature fields reconstruction methods~\cite{kobayashi2022decomposing, zhou2024feature,yue2024improving, fan2024large, zhou2025feature4x} have significantly enhanced the vision foundation model's performance in 3D/4D space by integrating structured latent scene representations into the model’s inference stage. Inspired by the promising results of feature lifting, we explore enhancing the InternVideo2-8B model~\cite{wang2024internvideo2} with spatiotemporal awareness by adopting the strategy proposed in Feature4X~\cite{zhou2025feature4x}, which constructs the VLM’s 2D feature space along the time dimension into a 4D feature field. To assess this approach, we evaluate performance on a subset of the \texttt{VLM4D} benchmark, specifically leveraging all 50 videos from the DAVIS 2016 dataset~\cite{perazzi2016benchmark}.
Our experimental evaluation compares inference performance across three distinct input modalities: original 2D videos; rendered novel global-view videos (which provide broader contextual information in 2D format); and reconstructed global feature fields (which implicitly incorporate 4D scene-level information during reasoning). Table~\ref{tab:4D_recon_results} reveals that reconstructed feature fields achieve the highest accuracy across both reasoning types. This success stems from two key advantages: the inherent structure of 4D representations and the ability of feature field inference to avoid the rendering artifacts of RGB reconstruction (global view video). 
However, the current approach requires per-scene optimization as a post-processing step, limiting its generalizability and making it computationally intensive.

\begin{table}[t]
    \centering
    \renewcommand{\arraystretch}{1.2}
    \begin{tabular}{l|c}
        \toprule
        \textbf{Model} & \textbf{MC} \\
        \midrule
        \rowcolor[HTML]{e1f0f5} \textit{Original Model} \phantom{-} & \phantom{-} \\
        Qwen 2VL (7B)  & 38.3 \\
        \hdashline
        Qwen 2.5VL (7B) & 43.4 \\
        \midrule
        \rowcolor[HTML]{e1f0f5} \textit{Finetuned Model} & \\
        Qwen 2VL (7B) (R) &  54.5 \\
        Qwen 2VL (7B) (S) & 42.5 \\
        Qwen 2VL (7B) (R+S) & \textbf{55.5} \\
        \hdashline
        Qwen 2.5VL (7B) (R) & 55.6 \\
        Qwen 2.5VL (7B) (S) & 48.6 \\
        Qwen 2.5VL (7B) (R+S) & \textbf{56.3} \\
        \bottomrule
    \end{tabular}
    \caption{\textbf{SFT on Spatial-Temporal Datasets.} MC refers to the performance of multiple-choice, while R, S, and R+S denote real, synthetic, and both real+synthetic usage of data for fine-tuning.}
    \label{tab:sft}
\end{table}

\begin{table}[t]
    \centering
    \renewcommand{\arraystretch}{1.2}
    \begin{tabular}{l|cc}
        \toprule
        \textbf{Input Modality} & \textbf{Accuracy} \\
        \midrule
        \rowcolor[HTML]{e1f0f5} \textit{Chain of Thought Response} & \phantom{-} & \phantom{-} \\
        Original 2D Video & 36.0 \\
        Global View Video & 32.7 \\
        Global Feature Field & \textbf{37.4} \\
        \midrule
        \rowcolor[HTML]{e1f0f5} \textit{Direct Output Response} & & \\
        Original 2D Video & 24.3 \\
        Global View Video & 23.8 \\
        Global Feature Field & \textbf{29.0} \\
        \bottomrule
    \end{tabular}
    \caption{\textbf{InternVideo2 Accuracy with 4D Feature Field Reconstruction.} Comparison of InternVideo2's performance given different input modalities from the same subset.}
    \label{tab:4D_recon_results}
    \vspace{-0.2cm}
\end{table}

%% file: sec/6_conclusion.tex
\section{Conclusion}
Through the construction of the~\texttt{VLM4D} benchmark, we evaluate the spatiotemporal reasoning capabilities of various vision language models (both open-source and proprietary). While more recently released models demonstrate improved performance over their counterparts, they remain significantly behind human proficiency. Overall, our work questions whether VLMs possess spatiotemporal reasoning abilities that are imperative to have for more sophisticated visual agents in fields ranging from robotics to interactive AI systems that require a deep understanding of dynamic visual environments. We hope to inspire future work to explore novel approaches for integrating spatiotemporal grounding, thereby enhancing their spatiotemporal reasoning capabilities and facilitating robust deployment. 
\newpage

%% file: sec/X_suppl.tex
\clearpage
\setcounter{section}{0}
\setcounter{figure}{0}
\setcounter{table}{0}
\maketitlesupplementary

\renewcommand\thesection{\Alph{section}} 
\renewcommand\thesubsection{\thesection.\arabic{subsection}} 
\renewcommand\thefigure{\Alph{figure}} 
\renewcommand\thetable{\Alph{table}} 

\crefname{section}{Sec.}{Secs.}
\Crefname{section}{Section}{Sections}
\Crefname{table}{Table}{Tables}
\crefname{table}{Tab.}{Tabs.}

\newcommand{\tabnohref}[1]{Tab.~{\color{red}#1}} 
\newcommand{\fignohref}[1]{Fig.~{\color{red}#1}} 
\newcommand{\secnohref}[1]{Sec.~{\color{red}#1}} 
\newcommand{\cnohref}[1]{[{\color{green}#1}]} 
\newcommand{\linenohref}[1]{Line~{\color{red}#1}}
\definecolor{iccvblue}{rgb}{0.21,0.49,0.74}

\noindent \textbf{Appendix Outline}
\begin{itemize}[itemsep=0em]
    \item Section~\ref{sec:vlm4d_stats}: Statistics of \texttt{VLM4D} benchmark
    \item Section~\ref{sec:eval_setup}: Evaluation setup 
    \item Section~\ref{sec:sft}: Analysis of Existing Video Instruction Tuning Datasets
    \item Section~\ref{sec:response examples}: Examples of VLMs’ responses with Chain-of-Thought and Direct Output prompting
    \item Section~\ref{sec:Details of Feature4x}: More details of 4D reconstruction using Feature4X
\end{itemize}


\section{VLM4D Benchmark Statistics}
\label{sec:vlm4d_stats}
\begin{table}[h]
    \centering
    \begin{tabular}{|l|c|c|}
        \hline
        \textbf{Dataset Statistics} & \textbf{Video} & \textbf{QA pair} \\
        \hline
        Real Samples & 600 & 1,371 \\
        Synthetic Samples & 400 & 445 \\
        Total Samples & 1,000 & 1,816 \\
        \hline
    \end{tabular}
    \caption{\textbf{VLM4D Dataset Breakdown}}
    \label{tab:vlm4d_stats}
\end{table}

\begin{figure*}[t]
  \centering
  \includegraphics[width=1\textwidth]{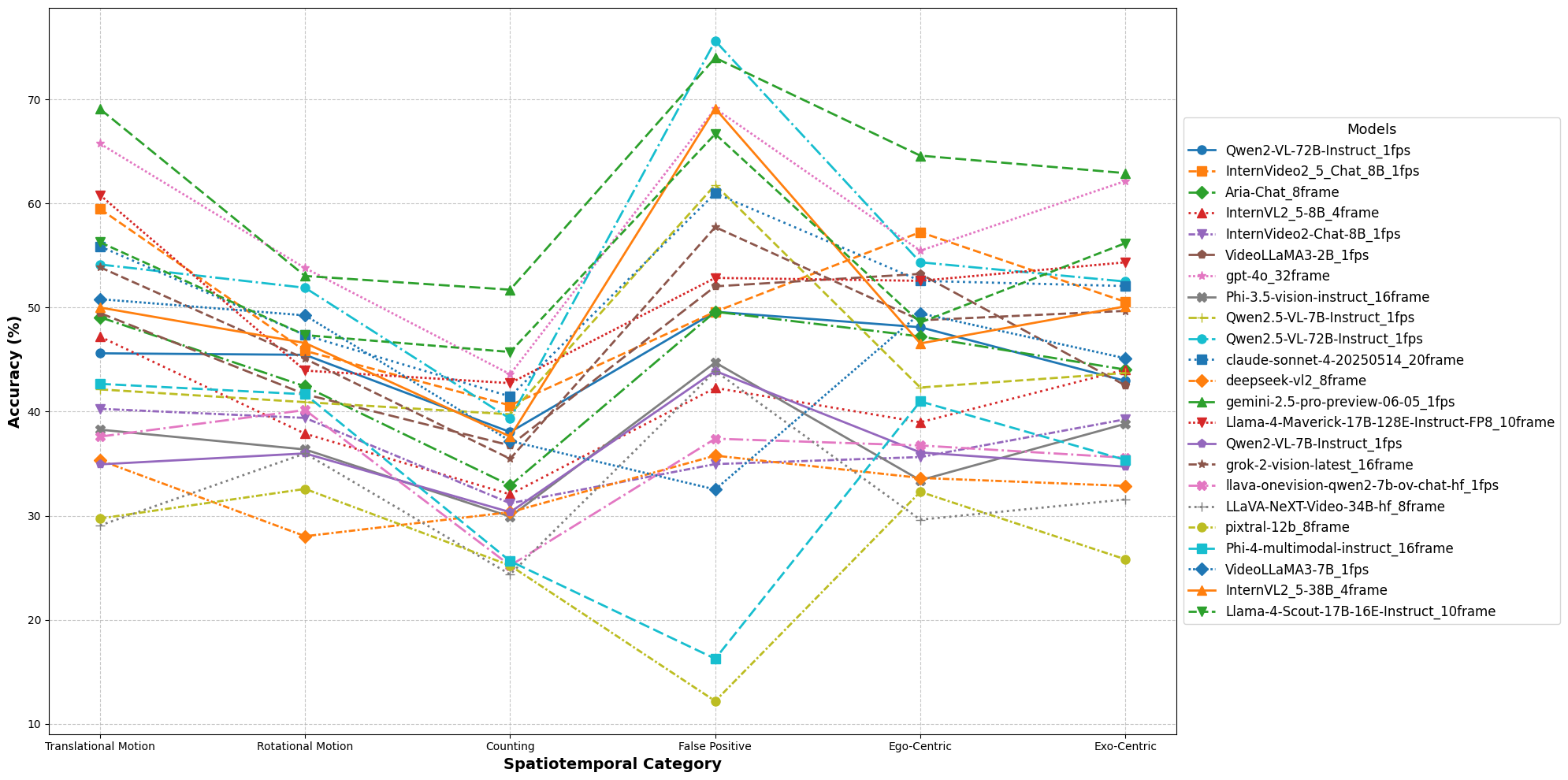}\vspace{-3mm}
  \caption{\textbf{Performance comparison of various VLMs across annotated
question categories including counting, rotational motion, translational motion, false positives, and action recognition}}
  \label{fig: model_acc_vs_category}
\end{figure*}

\cref{tab:vlm4d_stats} presents the breakdown of our \texttt{VLM4D} benchmark dataset. Additionally, \cref{fig: model_acc_vs_category} visualizes the detailed performance of VLMs across different question categories. For models that support only image input, we convert videos into multi-frame image sequences, using the maximum number of frames allowed within the model’s context window. For models that support video input, we follow their default frame rate settings, typically 1 fps.


\section{Evaluation setup}
\label{sec:eval_setup}
\begin{table}[h]
    \centering
    \begin{tabular}{|l|c|c|}
        \hline
        \textbf{Task} & \textbf{GPU Configuration} \\
        \hline
        Model Evaluation & 8 $\times$ A100 \\
        4D Feature Field Reconstruction & 1 $\times$ A100 \\
        \hline
    \end{tabular}
    \caption{\textbf{Evaluation Type and GPU Requirements}}
    \label{tab:evaluation_setup}
\end{table}


\section{Video Instruction Tuning Dataset Analysis}
\label{sec:sft}

We begin by analyzing four individual collections of datasets, which together contribute a substantial body of data for our experiments. The datasets used in this study are:

\begin{itemize}
    \item \textbf{ShareGPT-4o}
    \item \textbf{VideoChat2-IT}
    \item \textbf{ShareGPT4Video}
    \item \textbf{LLava-178k} 
\end{itemize}

As shown in~\cref{tab:dataset_sizes}, these datasets contain over 2 million samples in total, a robust foundation for evaluating and benchmarking the spatiotemporal validity of the highly used video instruction tuning datasets. 

\begin{figure*}[t]
  \centering
  \includegraphics{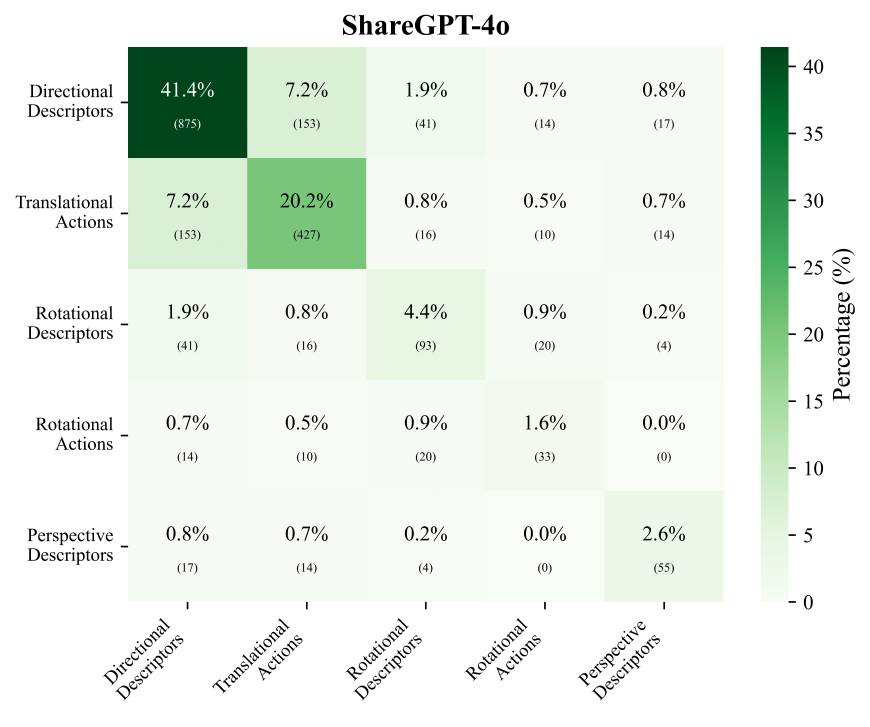}\vspace{-3mm}
  \caption{\textbf{Heatmap of Occurrences of Spatial-Temporal Terms in ShareGPT-4o}}
  \label{4o}
\end{figure*}

\begin{figure*}[t]
  \centering
  \includegraphics[width=0.9\textwidth]{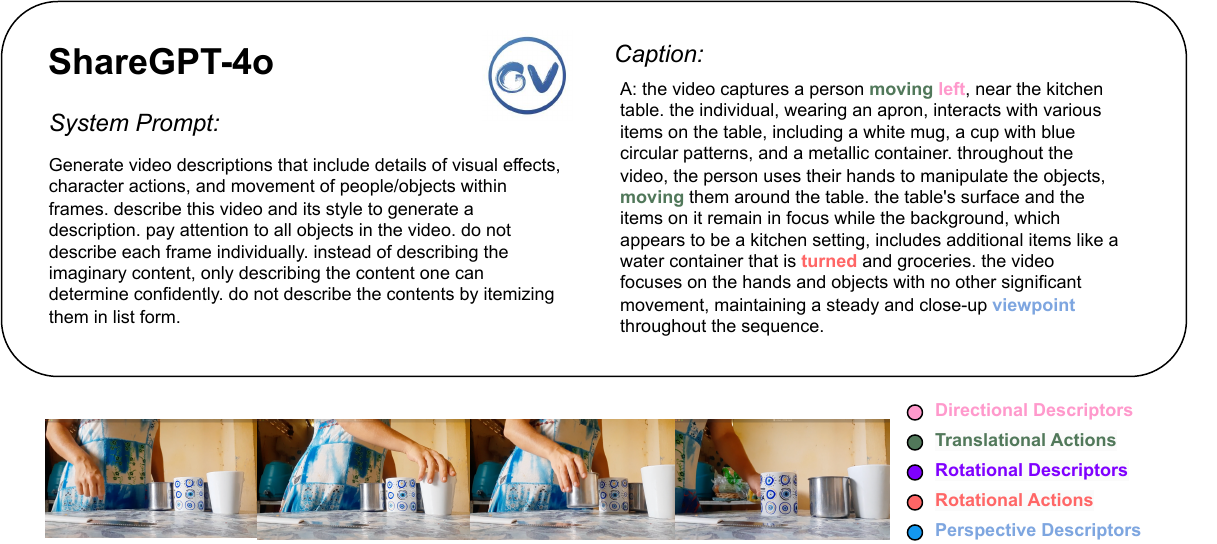}\hspace{8mm}  
    \caption{{\textit{Example of a clip with multiple target categories. Spatiotemporal grounding remains a challenge, as the direction descriptor is \textbf{incorrect}, and the translational actions provide limited insight, primarily indicating that objects/subjects are not static.}}}
  \label{ex}
\end{figure*}

\begin{figure*}[t]
  \centering
  \includegraphics[width=0.9\textwidth]{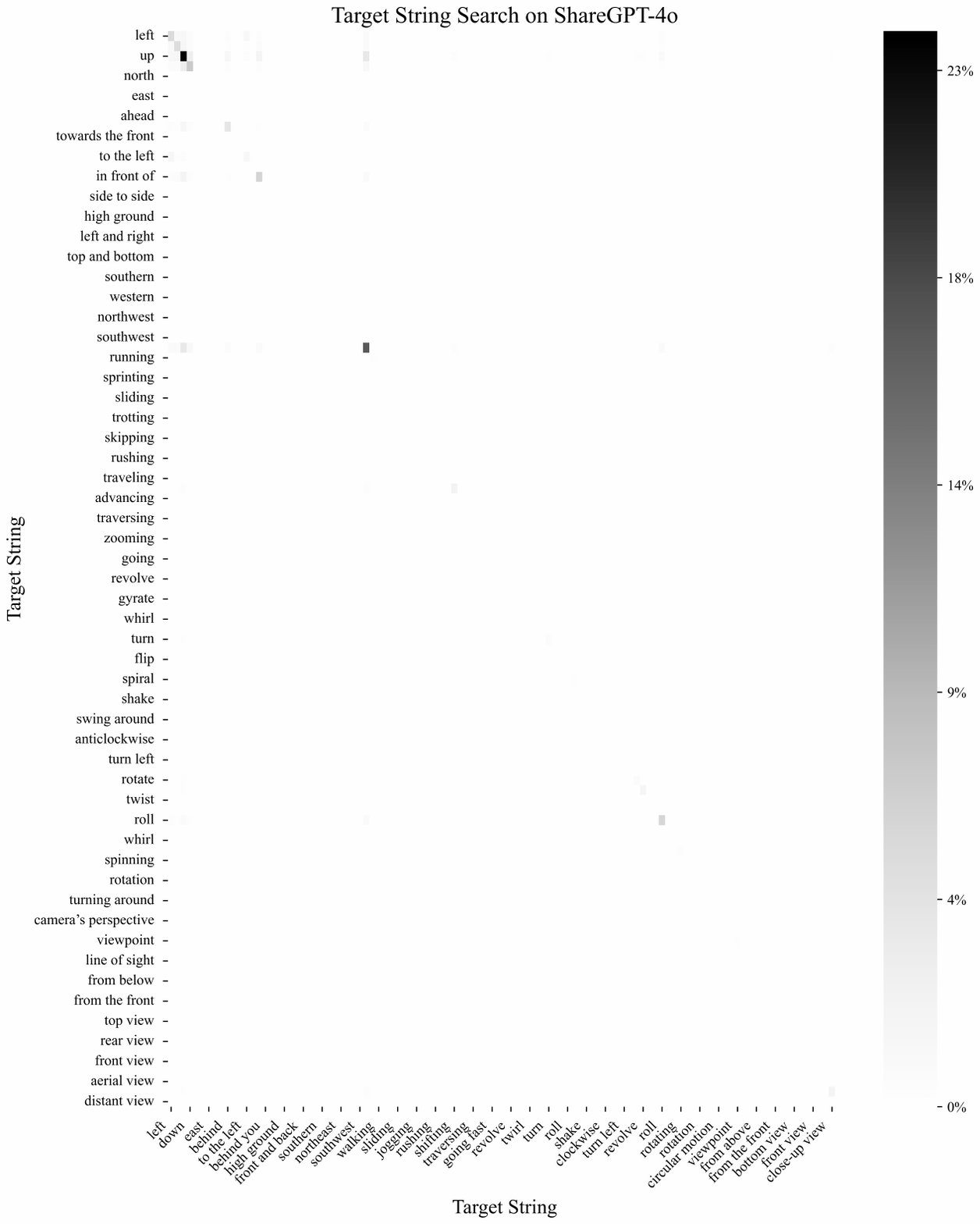}\hspace{8mm}  
    \caption{\textbf{Heatmap to visualize the prevalence of co-occurring target strings across the ShareGPT-4o dataset, informing our evaluation of spatiotemporal grounding in video instruction data.}}
  \label{TS}
\end{figure*}

We show our target strings in~\cref{tab:target_categories}. In our comprehensive analysis of the ShareGPT-4o dataset (\cref{4o}), we observed that over 40\% of the captions incorporate at least one target category. As depicted in~\cref{TS}, our target string search highlights a significant emphasis on the directional descriptors ``left" and ``right." Furthermore, the analysis of negative samples, illustrated in~\cref{ex}, indicates that these directional terms are seldom employed in conjunction with rotational or translational actions. This observation is further substantiated by the minimal overlaps between directional descriptors and action-related terms, as shown in~\cref{TS}. Collectively, this shows a notable gap in the dataset's ability to capture complex spatiotemporal relationships, particularly those involving dynamic textures and nuanced motion patterns. The statistics are summarized in~\cref{tab:caption_distribution}.

\begin{table}[h]
    \centering
    \renewcommand{\arraystretch}{1.2}
    \label{tab:dataset_distribution}
    \begin{tabular}{l c}
        \toprule
        \textbf{Dataset} & \textbf{Count} \\
        \midrule
        VideoChat2-IT & 423,497 \\
        ShareGPT4Video & 40,178 \\
        LLaVA-178K & 1,627,017 \\
        ShareGPT-4o & 2,111 \\
        \midrule
        \textbf{Total} & \textbf{2,092,803} \\
    \end{tabular}
     \caption{\textbf{No. of Video Instruction Tuning Samples (QA pair)}}
    \label{tab:dataset_sizes}
\end{table}

\begin{table}[h]
    \centering
    \begin{tabular}{|l|c|c|}
        \hline
        \textbf{Caption Category} & \textbf{Count} & \textbf{Percentage (\%)} \\
        \hline
        With Target String & 1,164,006 & 55.6 \\
        Without Target String & 928,897 & 44.4 \\
        \hline
    \end{tabular}
    \caption{\textbf{Distribution of Captions Containing Target Strings}}
    \label{tab:caption_distribution}
\end{table}

\begin{table*}[h]
    \centering
    \resizebox{1.0\linewidth}{!}{
    \renewcommand{\arraystretch}{1.2}
    \begin{tabular}{|l|p{15cm}|} 
    \hline
    \textbf{Category} & \textbf{Terms} \\
    \hline
    \textbf{Directional Descriptors} & left, right, up, down, north, south, east, west, ahead, behind, towards the front, away from the, to the left, toward, to the right, in front of, behind you, side to side, straight ahead, high ground, low ground, left and right, front and back, top and bottom, northern, southern, eastern, western, northeast, northwest, southeast, southwest, around \\
    \hline
    \textbf{Translational Actions} & moving, running, walking, sprinting, gliding, sliding, crawling, trotting, jogging, skipping, bounding, rushing, hurrying, traveling, shifting, advancing, progressing, traversing, racing, zooming, going fast, going \\
    \hline
    \textbf{Rotational Descriptors} & rotate, revolve, spin, gyrate, twirl, whirl, twist, turn, pivot, flip, roll, spiral, swing, shake, oscillate, swing around, rolls, roll \\
    \hline
    \textbf{Rotational Actions} & clockwise, anticlockwise, turn right, turn left, spin, rotate, revolve, twist, pivot, gyrate, whirl, rotating, spinning, turning 360 degrees, turning 180 degrees,  rotation, twisting, turning around, circular motion, turning 90 degrees, \\
    \hline
    \textbf{Perspective Descriptors} & camera’s perspective, frame’s perspective, viewpoint, point of view, line of sight, from above, from below, from the side, from the front, from behind, top view, bottom view, rear view, side view, front view, bird’s eye view, aerial view, close-up view, distant view \\
    \hline
    \end{tabular}
    }
    \caption{\textbf{Video Instruction Tuning Datasets Categories \& Target strings Analysis}}
\label{tab:target_categories}
\end{table*}

\begin{figure*}[t]
  \centering
  \includegraphics{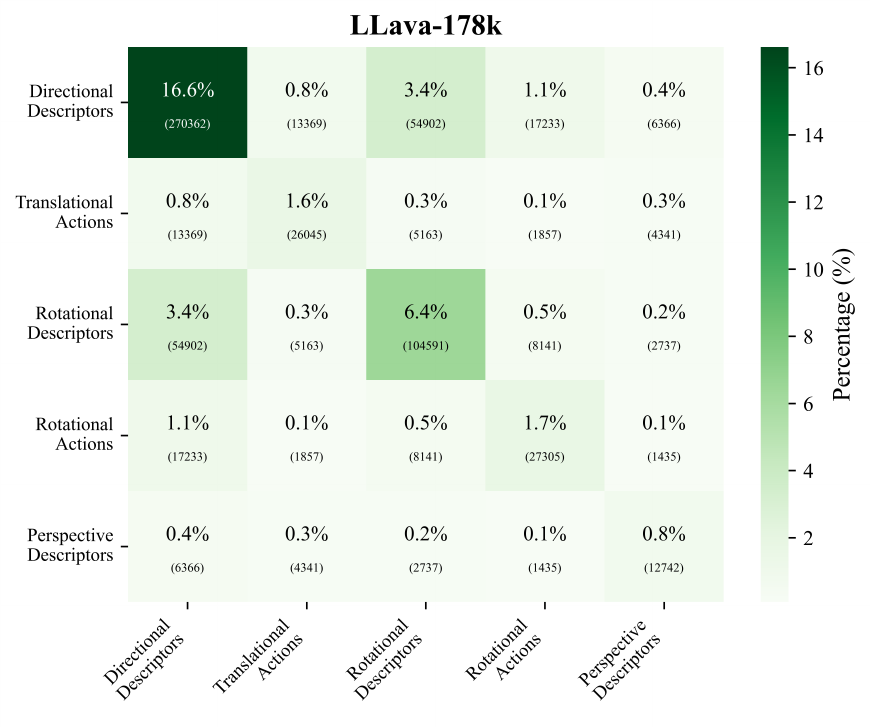}\vspace{-3mm}
  \caption{\textbf{Heatmap of Occurrences of Spatial-Temporal Terms in LLava-178k}}
  
    {\small \textit{(Note: LLava-178k actually comprises over 1.6 million samples, as we combine many of the available dataset splits within this collection.)}}
    
  \label{Q&A_example1}
\end{figure*}

\begin{figure*}[t]
  \centering
  \includegraphics{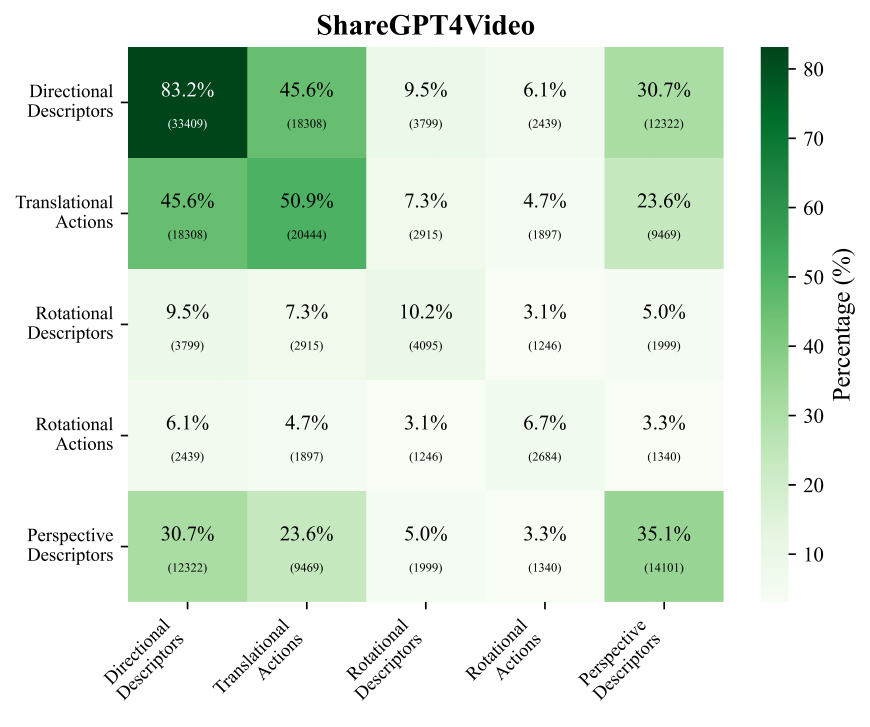}\vspace{-3mm}
  \caption{\textbf{Heatmap of Occurrences of Spatial-Temporal Terms in ShareGPT4Video}}
  \label{Q&A_example2}
\end{figure*}

\begin{figure*}[t]
  \centering
  \includegraphics{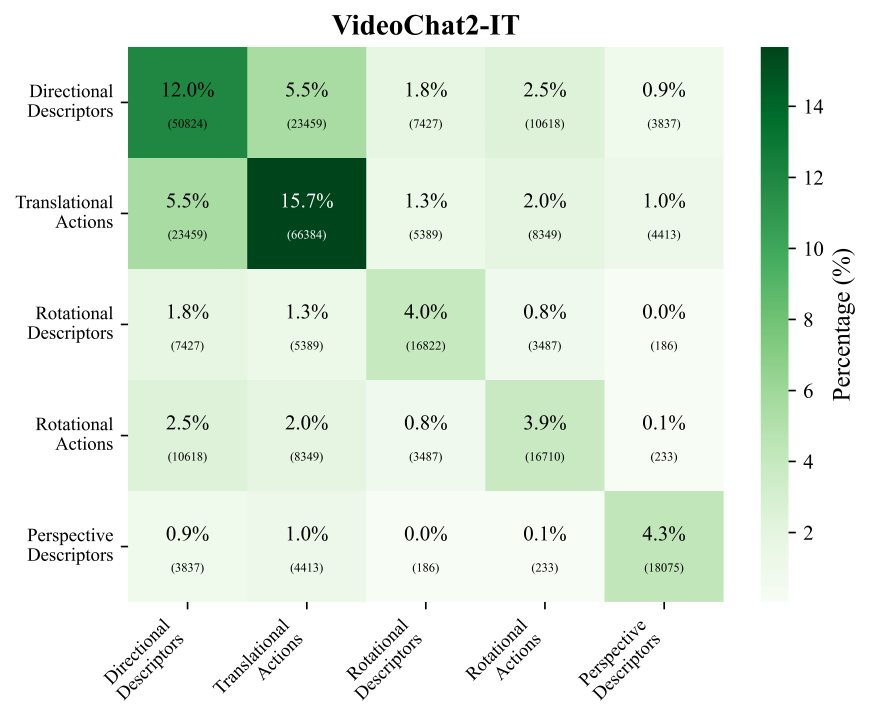}\vspace{-3mm}
  \caption{\textbf{Heatmap of Occurrences of Spatial-Temporal Terms in VideoChat-IT}}
  \label{Q&A_example3}
\end{figure*}

\section{Examples of VLMs' responses}
\label{sec:response examples}
\begin{figure*}[t]
  \centering
  \includegraphics[width=0.9\textwidth]{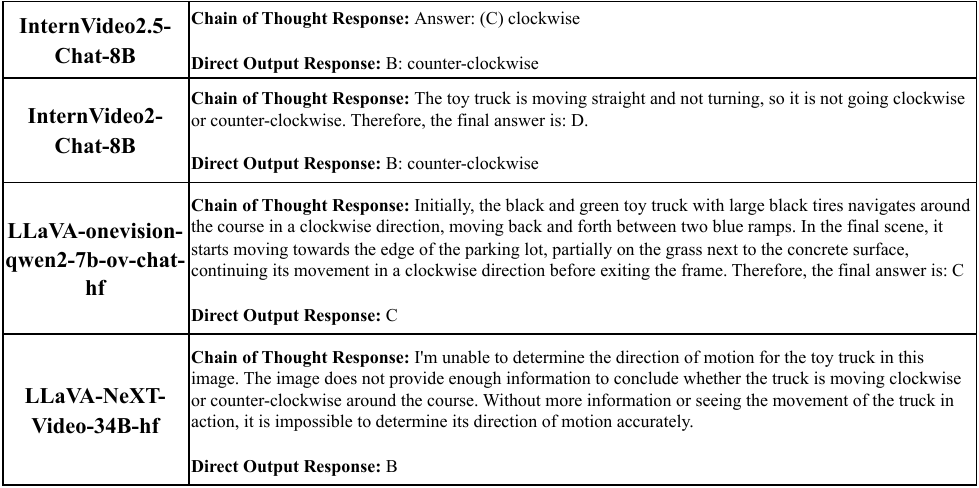}\vspace{-3mm}
  \caption{Models 20-23}
  \label{fig:Q&A_example_5}
\end{figure*}

Please refer to Fig. \textcolor{iccvblue}{I} - \ref{fig:Q&A_example_5} for detailed responses from all evaluated VLMs under both chain-of-thought (CoT) and direct output (DO) prompting, based on the given example video and question.

\section{Details of 4D Reconstruction}
\label{sec:Details of Feature4x}
We utilize the Feature4X framework (Fig. \textcolor{iccvblue}{H}) for 4D reconstruction experiments conducted on our dataset. Given an input monocular RGB video, Feature4X reconstructs the dynamic 3D scene by employing dynamic 3D Gaussians, specifically Dynamic 3D Gaussian Splatting, which represent dynamic foreground elements that deform over time. These dynamic Gaussians are guided by a 4D Motion Scaffold, a sparse graph of trajectory nodes, enabling the interpolation of dense motion trajectories and features for each Gaussian efficiently. A separate set of static 3D Gaussians represents static background elements.

Feature4X introduces a unified latent feature embedding, distilled from various foundational 2D models, which facilitates multiple downstream tasks such as segmentation, scene editing, and visual question answering (VQA). Specifically, Feature4X extracts video segment features from the InternVideo2-Chat model, a foundation model fine-tuned for video question answering. 

This unified feature field is directly used by the InternVideo decoder for inference, bypassing the video encoding step entirely. This approach significantly improves inference efficiency and retains comprehensive structural information from the 4D scene representation, which surpasses the context available from the original 2D videos alone. Consequently, this method enhances downstream tasks by providing richer spatiotemporal context and semantic consistency.

\begin{figure*}[t]
  \centering
  \includegraphics[width=1\textwidth]{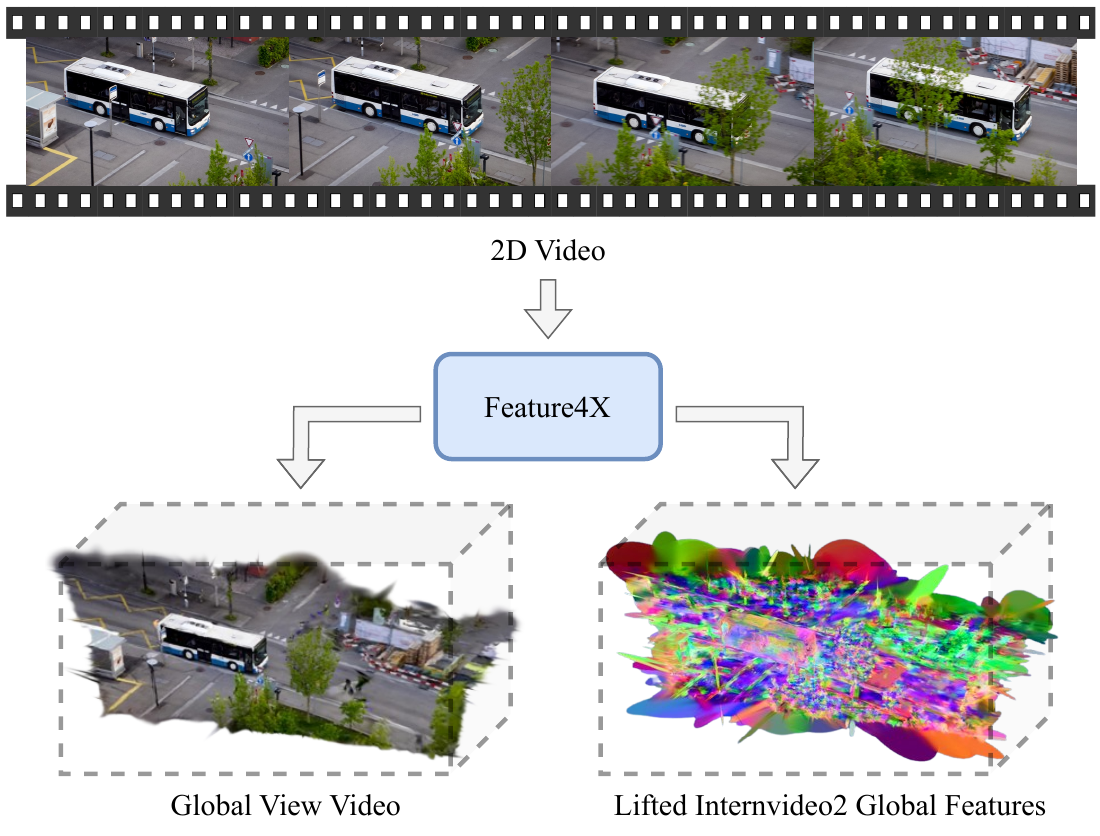}\vspace{-3mm}
  \label{fig:4D_recon}
  \caption{\textbf{General Input-Output Architecture of Feature4X} }
\end{figure*}

\begin{figure*}[t]
  \centering
  \includegraphics[width=0.9\textwidth]{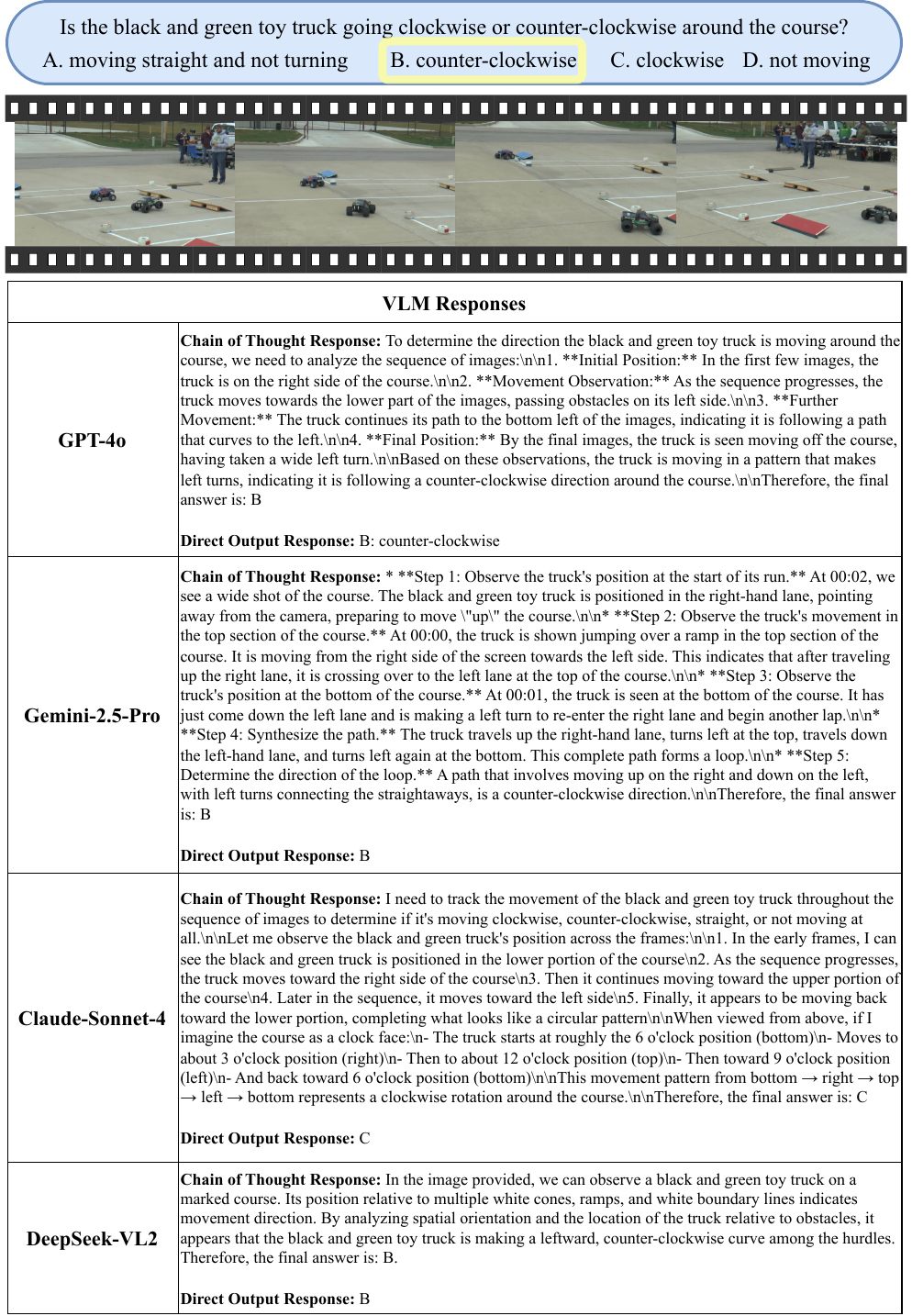}\vspace{-3mm}
  \label{fig:Q&A_example_1}
  \caption{\textbf{Complete CoT and DO Responses of VLMs} Models 1-4}
\end{figure*}

\begin{figure*}[t]
  \centering
  \includegraphics[width=0.9\textwidth]{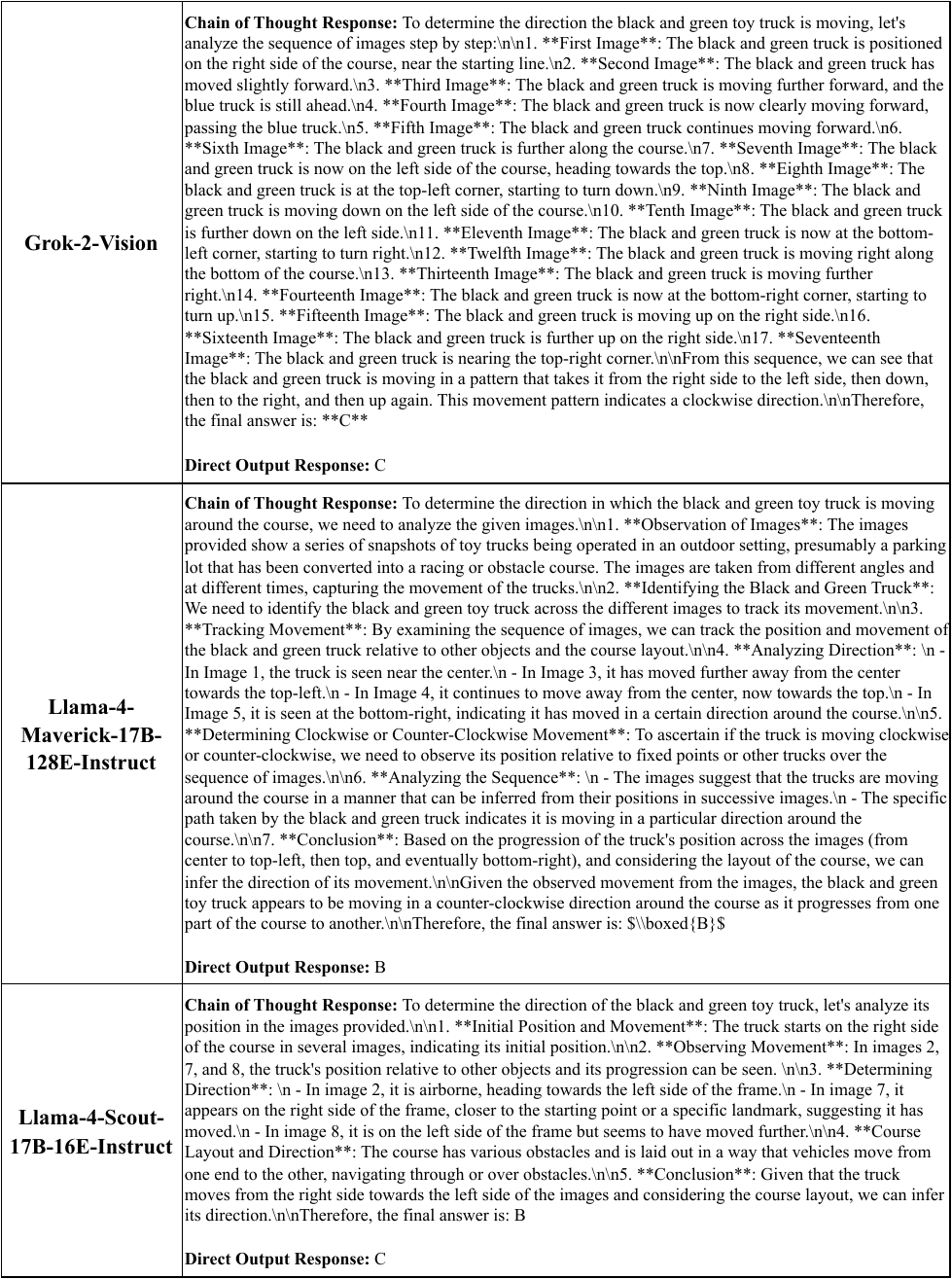}\vspace{-3mm}
  \caption{Models 5-7}
  \label{fig:Q&A_example_2}
\end{figure*}

\begin{figure*}[t]
  \centering
  \includegraphics[width=0.9\textwidth]{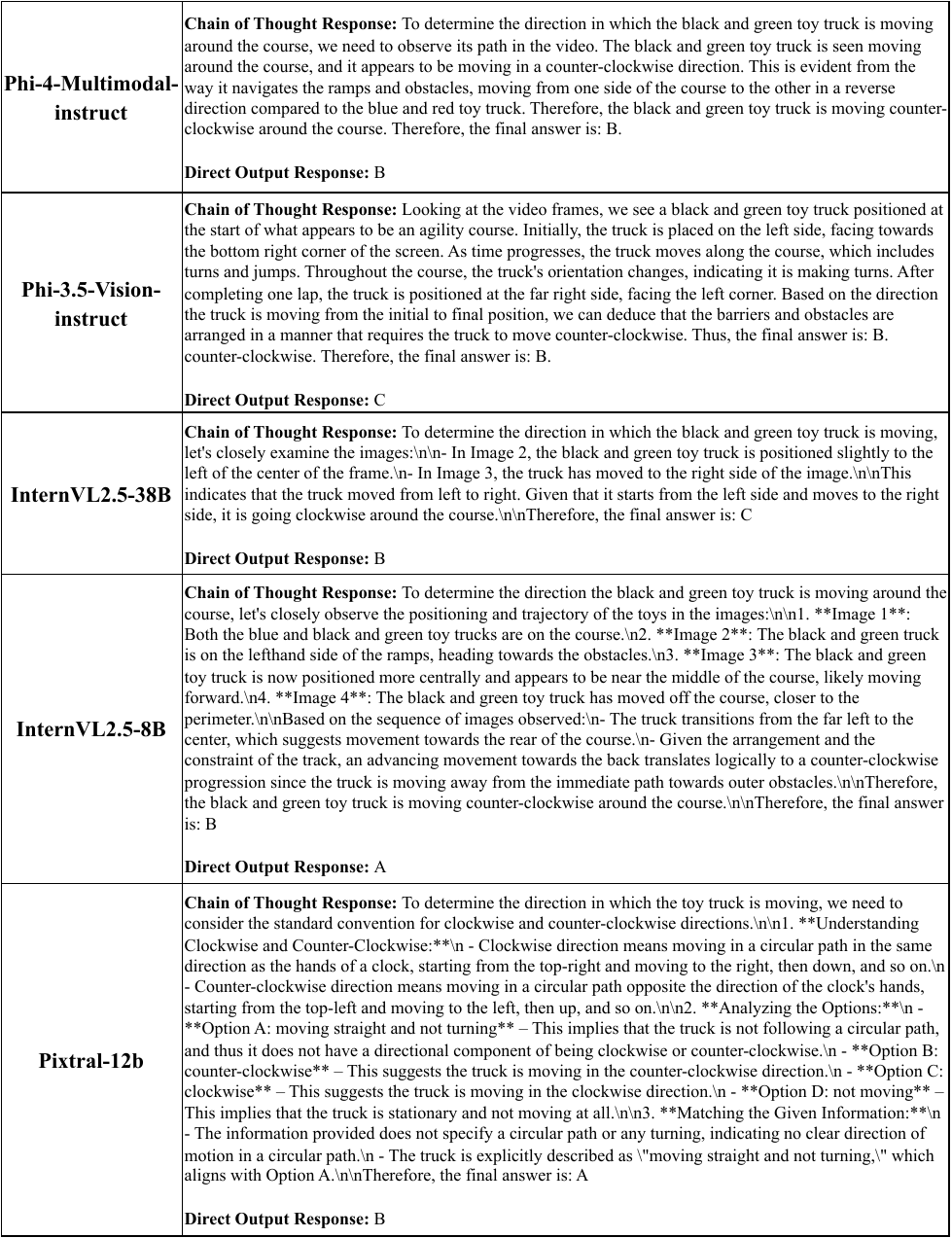}\vspace{-3mm}
  \caption{Models 8-12}
  \label{fig:Q&A_example_3}
\end{figure*}

\begin{figure*}[t]
  \centering
  \includegraphics[width=0.9\textwidth]{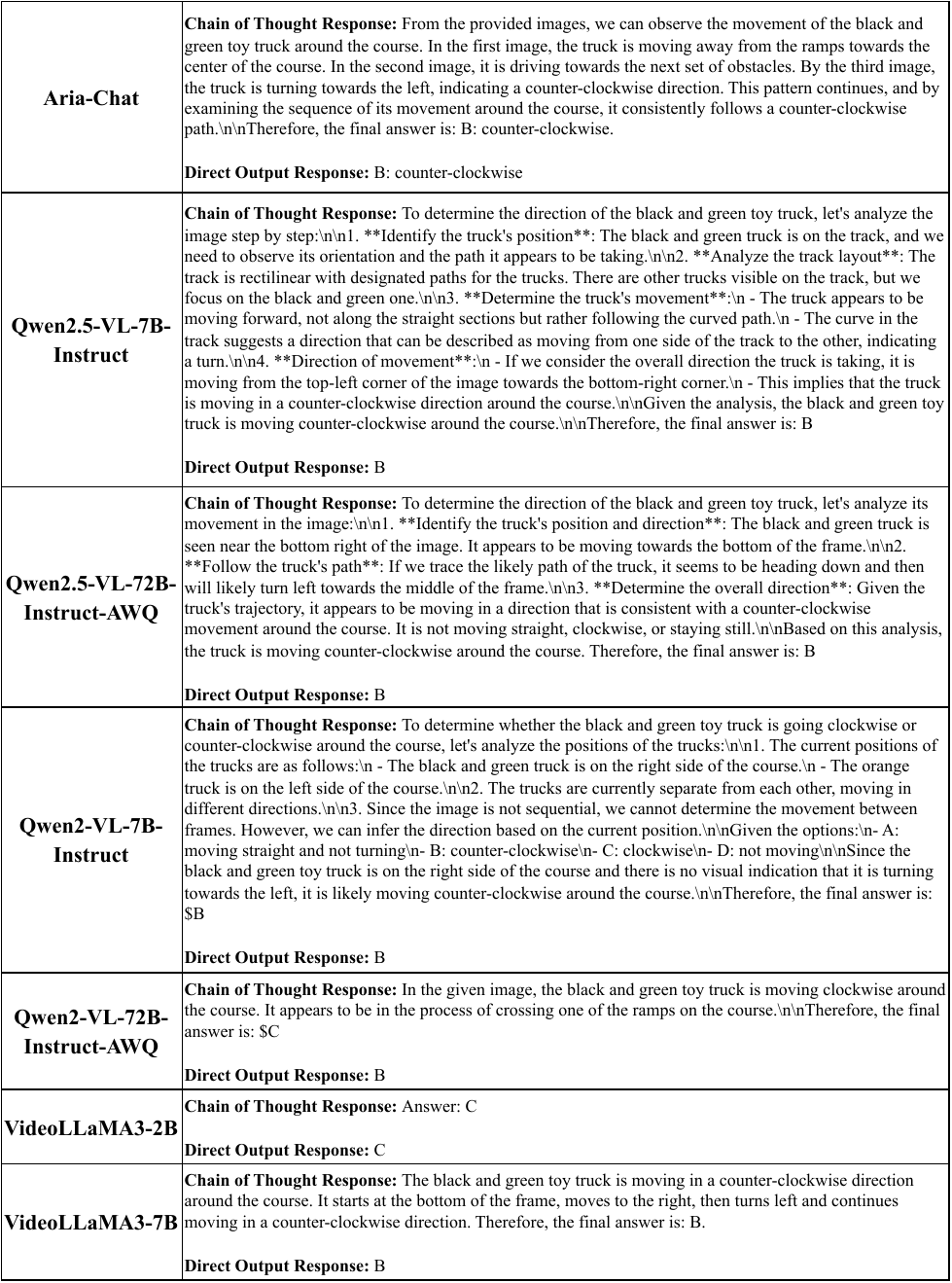}\vspace{-3mm}
  \caption{Models 13-19}
  \label{fig:Q&A_example_4}
\end{figure*}

%



%% file: main.bbl
\begin{thebibliography}{104}
\providecommand{\natexlab}[1]{#1}
\providecommand{\url}[1]{\texttt{#1}}
\expandafter\ifx\csname urlstyle\endcsname\relax
  \providecommand{\doi}[1]{doi: #1}\else
  \providecommand{\doi}{doi: \begingroup \urlstyle{rm}\Url}\fi

\bibitem[Abdin et~al.(2024{\natexlab{a}})Abdin, Aneja, Awadalla, Awadallah, Awan, Bach, Bahree, Bakhtiari, Bao, Behl, et~al.]{abdin2024phi3}
Marah Abdin, Jyoti Aneja, Hany Awadalla, Ahmed Awadallah, Ammar~Ahmad Awan, Nguyen Bach, Amit Bahree, Arash Bakhtiari, Jianmin Bao, Harkirat Behl, et~al.
\newblock Phi-3 technical report: A highly capable language model locally on your phone.
\newblock \emph{arXiv preprint arXiv:2404.14219}, 2024{\natexlab{a}}.

\bibitem[Abdin et~al.(2024{\natexlab{b}})Abdin, Aneja, Behl, Bubeck, Eldan, Gunasekar, Harrison, Hewett, Javaheripi, Kauffmann, et~al.]{abdin2024phi4}
Marah Abdin, Jyoti Aneja, Harkirat Behl, S{\'e}bastien Bubeck, Ronen Eldan, Suriya Gunasekar, Michael Harrison, Russell~J Hewett, Mojan Javaheripi, Piero Kauffmann, et~al.
\newblock Phi-4 technical report.
\newblock \emph{arXiv preprint arXiv:2412.08905}, 2024{\natexlab{b}}.

\bibitem[Abouelenin et~al.(2025)Abouelenin, Ashfaq, Atkinson, Awadalla, Bach, Bao, Benhaim, Cai, Chaudhary, Chen, Chen, Chen, Chen, Chen, Chen, ling Chen, Dai, Dai, Fan, Gao, Gao, Garg, Goswami, Hao, Hendy, Hu, Jin, Khademi, Kim, Kim, Lee, Li, Li, Liang, Lin, Lin, Liu, Liu, Lopez, Luo, Madan, Mazalov, Mousavi, Nguyen, Pan, Perez-Becker, Platin, Portet, Qiu, Ren, Ren, Roy, Shang, Shen, Singhal, Som, Song, Sych, Vaddamanu, Wang, Wang, Wang, Wu, Xu, Xu, Yang, Yang, Yu, Zabir, Zhang, Zhang, Zhang, and Zhou]{abouelenin2025phi4minitechnicalreportcompact}
Abdelrahman Abouelenin, Atabak Ashfaq, Adam Atkinson, Hany Awadalla, Nguyen Bach, Jianmin Bao, Alon Benhaim, Martin Cai, Vishrav Chaudhary, Congcong Chen, Dong Chen, Dongdong Chen, Junkun Chen, Weizhu Chen, Yen-Chun Chen, Yi ling Chen, Qi Dai, Xiyang Dai, Ruchao Fan, Mei Gao, Min Gao, Amit Garg, Abhishek Goswami, Junheng Hao, Amr Hendy, Yuxuan Hu, Xin Jin, Mahmoud Khademi, Dongwoo Kim, Young~Jin Kim, Gina Lee, Jinyu Li, Yunsheng Li, Chen Liang, Xihui Lin, Zeqi Lin, Mengchen Liu, Yang Liu, Gilsinia Lopez, Chong Luo, Piyush Madan, Vadim Mazalov, Ali Mousavi, Anh Nguyen, Jing Pan, Daniel Perez-Becker, Jacob Platin, Thomas Portet, Kai Qiu, Bo Ren, Liliang Ren, Sambuddha Roy, Ning Shang, Yelong Shen, Saksham Singhal, Subhojit Som, Xia Song, Tetyana Sych, Praneetha Vaddamanu, Shuohang Wang, Yiming Wang, Zhenghao Wang, Haibin Wu, Haoran Xu, Weijian Xu, Yifan Yang, Ziyi Yang, Donghan Yu, Ishmam Zabir, Jianwen Zhang, Li~Lyna Zhang, Yunan Zhang, and Xiren Zhou.
\newblock Phi-4-mini technical report: Compact yet powerful multimodal language models via mixture-of-loras, 2025.

\bibitem[Agarwal et~al.(2025)Agarwal, Ali, Bala, Balaji, Barker, Cai, Chattopadhyay, Chen, Cui, Ding, et~al.]{agarwal2025cosmos}
Niket Agarwal, Arslan Ali, Maciej Bala, Yogesh Balaji, Erik Barker, Tiffany Cai, Prithvijit Chattopadhyay, Yongxin Chen, Yin Cui, Yifan Ding, et~al.
\newblock Cosmos world foundation model platform for physical ai.
\newblock \emph{arXiv preprint arXiv:2501.03575}, 2025.

\bibitem[Agrawal et~al.(2024)Agrawal, Antoniak, Hanna, Bout, Chaplot, Chudnovsky, Costa, De~Monicault, Garg, Gervet, et~al.]{agrawal2024pixtral}
Pravesh Agrawal, Szymon Antoniak, Emma~Bou Hanna, Baptiste Bout, Devendra Chaplot, Jessica Chudnovsky, Diogo Costa, Baudouin De~Monicault, Saurabh Garg, Theophile Gervet, et~al.
\newblock Pixtral 12b.
\newblock \emph{arXiv preprint arXiv:2410.07073}, 2024.

\bibitem[Anthropic(2025)]{claude4}
Anthropic.
\newblock System card: Claude opus 4 \& claude sonnet 4.
\newblock Technical report, 2025.

\bibitem[Bai et~al.(2023)Bai, Bai, Chu, Cui, Dang, Deng, Fan, Ge, Han, Huang, et~al.]{bai2023qwen}
Jinze Bai, Shuai Bai, Yunfei Chu, Zeyu Cui, Kai Dang, Xiaodong Deng, Yang Fan, Wenbin Ge, Yu Han, Fei Huang, et~al.
\newblock Qwen technical report.
\newblock \emph{arXiv preprint arXiv:2309.16609}, 2023.

\bibitem[Brooks et~al.(2024)Brooks, Peebles, Holmes, DePue, Guo, Jing, Schnurr, Taylor, Luhman, Luhman, Ng, Wang, and Ramesh]{videoworldsimulators2024}
Tim Brooks, Bill Peebles, Connor Holmes, Will DePue, Yufei Guo, Li Jing, David Schnurr, Joe Taylor, Troy Luhman, Eric Luhman, Clarence Ng, Ricky Wang, and Aditya Ramesh.
\newblock Video generation models as world simulators.
\newblock 2024.

\bibitem[Brown et~al.(2020)Brown, Mann, Ryder, Subbiah, Kaplan, Dhariwal, Neelakantan, Shyam, Sastry, Askell, et~al.]{brown2020language}
Tom Brown, Benjamin Mann, Nick Ryder, Melanie Subbiah, Jared~D Kaplan, Prafulla Dhariwal, Arvind Neelakantan, Pranav Shyam, Girish Sastry, Amanda Askell, et~al.
\newblock Language models are few-shot learners.
\newblock \emph{Advances in neural information processing systems}, 33:\penalty0 1877--1901, 2020.

\bibitem[Burgess(2006)]{Burgess2006Spatial}
Neil Burgess.
\newblock Spatial memory: how egocentric and allocentric combine.
\newblock \emph{Trends in Cognitive Sciences}, 10\penalty0 (12):\penalty0 551--557, 2006.

\bibitem[Chari et~al.(2023)Chari, Ba, Zhou, Talegaonkar, Athreya, and Kadambi]{chari2023learning}
Pradyumna Chari, Yunhao Ba, Shijie Zhou, Chinmay Talegaonkar, Shreeram Athreya, and Achuta Kadambi.
\newblock On learning mechanical laws of motion from video using neural networks.
\newblock \emph{IEEE Access}, 11:\penalty0 30129--30145, 2023.

\bibitem[Chen et~al.(2024{\natexlab{a}})Chen, Xu, Kirmani, Ichter, Sadigh, Guibas, and Xia]{chen2024spatialvlm}
Boyuan Chen, Zhuo Xu, Sean Kirmani, Brain Ichter, Dorsa Sadigh, Leonidas Guibas, and Fei Xia.
\newblock Spatialvlm: Endowing vision-language models with spatial reasoning capabilities.
\newblock In \emph{Proceedings of the IEEE/CVF Conference on Computer Vision and Pattern Recognition}, pages 14455--14465, 2024{\natexlab{a}}.

\bibitem[Chen et~al.(2022)Chen, Guo, Yi, Li, and Elhoseiny]{Chen_2022_CVPR}
Jun Chen, Han Guo, Kai Yi, Boyang Li, and Mohamed Elhoseiny.
\newblock Visualgpt: Data-efficient adaptation of pretrained language models for image captioning.
\newblock In \emph{Proceedings of the IEEE/CVF Conference on Computer Vision and Pattern Recognition (CVPR)}, pages 18030--18040, 2022.

\bibitem[Chen et~al.(2024{\natexlab{b}})Chen, Li, Dong, Zhang, Zang, Chen, Duan, Wang, Qiao, Lin, et~al.]{chen2024we}
Lin Chen, Jinsong Li, Xiaoyi Dong, Pan Zhang, Yuhang Zang, Zehui Chen, Haodong Duan, Jiaqi Wang, Yu Qiao, Dahua Lin, et~al.
\newblock Are we on the right way for evaluating large vision-language models?
\newblock \emph{arXiv preprint arXiv:2403.20330}, 2024{\natexlab{b}}.

\bibitem[Chen et~al.(2024{\natexlab{c}})Chen, Wei, Li, Dong, Zhang, Zang, Chen, Duan, Lin, Tang, et~al.]{chen2024sharegpt4video}
Lin Chen, Xilin Wei, Jinsong Li, Xiaoyi Dong, Pan Zhang, Yuhang Zang, Zehui Chen, Haodong Duan, Bin Lin, Zhenyu Tang, et~al.
\newblock Sharegpt4video: Improving video understanding and generation with better captions.
\newblock \emph{arXiv preprint arXiv:2406.04325}, 2024{\natexlab{c}}.

\bibitem[Chen et~al.(2024{\natexlab{d}})Chen, Wang, Cao, Liu, Gao, Cui, Zhu, Ye, Tian, Liu, et~al.]{chen2024expanding}
Zhe Chen, Weiyun Wang, Yue Cao, Yangzhou Liu, Zhangwei Gao, Erfei Cui, Jinguo Zhu, Shenglong Ye, Hao Tian, Zhaoyang Liu, et~al.
\newblock Expanding performance boundaries of open-source multimodal models with model, data, and test-time scaling.
\newblock \emph{arXiv preprint arXiv:2412.05271}, 2024{\natexlab{d}}.

\bibitem[Cheng et~al.(2024{\natexlab{a}})Cheng, Yin, Fu, Guo, Yang, Kautz, Wang, and Liu]{cheng2024spatialrgpt}
An-Chieh Cheng, Hongxu Yin, Yang Fu, Qiushan Guo, Ruihan Yang, Jan Kautz, Xiaolong Wang, and Sifei Liu.
\newblock Spatialrgpt: Grounded spatial reasoning in vision-language models.
\newblock \emph{Advances in Neural Information Processing Systems}, 37:\penalty0 135062--135093, 2024{\natexlab{a}}.

\bibitem[Cheng et~al.(2024{\natexlab{b}})Cheng, Leng, Zhang, Xin, Li, Chen, Zhu, Zhang, Luo, Zhao, et~al.]{cheng2024videollama}
Zesen Cheng, Sicong Leng, Hang Zhang, Yifei Xin, Xin Li, Guanzheng Chen, Yongxin Zhu, Wenqi Zhang, Ziyang Luo, Deli Zhao, et~al.
\newblock Videollama 2: Advancing spatial-temporal modeling and audio understanding in video-llms.
\newblock \emph{arXiv preprint arXiv:2406.07476}, 2024{\natexlab{b}}.

\bibitem[Cui et~al.(2024)Cui, He, Ma, Chen, Tian, Wang, Li, Wang, Wang, Zhu, Lu, Lu, Wang, Wang, Qiao, and Dai]{cui2025comprehensive}
Erfei Cui, Yinan He, Zheng Ma, Zhe Chen, Hao Tian, Weiyun Wang, Kunchang Li, Yi Wang, Wenhai Wang, Xizhou Zhu, Lewei Lu, Tong Lu, Yali Wang, Limin Wang, Yu Qiao, and Jifeng Dai.
\newblock Sharegpt-4o: Comprehensive multimodal annotations with gpt-4o, 2024.

\bibitem[Dai et~al.(2023)Dai, Li, Li, Tiong, Zhao, Wang, Li, Fung, and Hoi]{dai2023instructblip}
Wenliang Dai, Junnan Li, Dongxu Li, Anthony Meng~Huat Tiong, Junqi Zhao, Weisheng Wang, Boyang Li, Pascale Fung, and Steven Hoi.
\newblock Instructblip: Towards general-purpose vision-language models with instruction tuning, 2023.

\bibitem[De~Freitas et~al.(2016)De~Freitas, Myers, and Nobre]{DeFreitas2016Tracking}
Julian De~Freitas, Nicholas~E. Myers, and Anna~C. Nobre.
\newblock Tracking the changing feature of a moving object.
\newblock \emph{Journal of Vision}, 16\penalty0 (3):\penalty0 22, 2016.

\bibitem[Devlin et~al.(2019)Devlin, Chang, Lee, and Toutanova]{devlin2019bert}
Jacob Devlin, Ming-Wei Chang, Kenton Lee, and Kristina Toutanova.
\newblock Bert: Pre-training of deep bidirectional transformers for language understanding.
\newblock In \emph{Proceedings of the 2019 conference of the North American chapter of the association for computational linguistics: human language technologies, volume 1 (long and short papers)}, pages 4171--4186, 2019.

\bibitem[Dosovitskiy et~al.(2021)Dosovitskiy, Beyer, Kolesnikov, Weissenborn, Zhai, Unterthiner, Dehghani, Minderer, Heigold, Gelly, Uszkoreit, and Houlsby]{Dosovitskiy2021An}
Alexey Dosovitskiy, Lucas Beyer, Alexander Kolesnikov, Dirk Weissenborn, Xiaohua Zhai, Thomas Unterthiner, Mostafa Dehghani, Matthias Minderer, Georg Heigold, Sylvain Gelly, Jakob Uszkoreit, and Neil Houlsby.
\newblock An image is worth 16x16 words: Transformers for image recognition at scale.
\newblock In \emph{International Conference on Learning Representations}, 2021.

\bibitem[Driess et~al.(2023)Driess, Xia, Sajjadi, Lynch, Chowdhery, Wahid, Tompson, Vuong, Yu, Huang, et~al.]{driess2023palm}
Danny Driess, Fei Xia, Mehdi~SM Sajjadi, Corey Lynch, Aakanksha Chowdhery, Ayzaan Wahid, Jonathan Tompson, Quan Vuong, Tianhe Yu, Wenlong Huang, et~al.
\newblock Palm-e: An embodied multimodal language model.
\newblock 2023.

\bibitem[Fan et~al.(2025{\natexlab{a}})Fan, Li, Sun, and Zhou]{fan2025missing}
Chenrui Fan, Ming Li, Lichao Sun, and Tianyi Zhou.
\newblock Missing premise exacerbates overthinking: Are reasoning models losing critical thinking skill?
\newblock \emph{arXiv preprint arXiv:2504.06514}, 2025{\natexlab{a}}.

\bibitem[Fan et~al.(2024)Fan, Zhang, Cong, Wang, Li, Wen, Zhou, Kadambi, Wang, Xu, et~al.]{fan2024large}
Zhiwen Fan, Jian Zhang, Wenyan Cong, Peihao Wang, Renjie Li, Kairun Wen, Shijie Zhou, Achuta Kadambi, Zhangyang Wang, Danfei Xu, et~al.
\newblock Large spatial model: End-to-end unposed images to semantic 3d.
\newblock \emph{Advances in neural information processing systems}, 37:\penalty0 40212--40229, 2024.

\bibitem[Fan et~al.(2025{\natexlab{b}})Fan, Zhang, Li, Zhang, Chen, Hu, Wang, Qu, Wang, Yan, et~al.]{fan2025vlm}
Zhiwen Fan, Jian Zhang, Renjie Li, Junge Zhang, Runjin Chen, Hezhen Hu, Kevin Wang, Huaizhi Qu, Dilin Wang, Zhicheng Yan, et~al.
\newblock Vlm-3r: Vision-language models augmented with instruction-aligned 3d reconstruction.
\newblock \emph{arXiv preprint arXiv:2505.20279}, 2025{\natexlab{b}}.

\bibitem[Freyd and Finke(1984)]{Freyd1984Representational}
Jennifer~J. Freyd and Ronald~A. Finke.
\newblock Representational momentum.
\newblock \emph{Journal of Experimental Psychology: Learning, Memory, and Cognition}, 10\penalty0 (1):\penalty0 126--132, 1984.

\bibitem[Fu et~al.(2024)Fu, Dai, Luo, Li, Ren, Zhang, Wang, Zhou, Shen, Zhang, et~al.]{fu2024video}
Chaoyou Fu, Yuhan Dai, Yongdong Luo, Lei Li, Shuhuai Ren, Renrui Zhang, Zihan Wang, Chenyu Zhou, Yunhang Shen, Mengdan Zhang, et~al.
\newblock Video-mme: The first-ever comprehensive evaluation benchmark of multi-modal llms in video analysis.
\newblock \emph{arXiv preprint arXiv:2405.21075}, 2024.

\bibitem[Gemini~Team(2025)]{gemini2_5}
Google Gemini~Team.
\newblock Gemini 2.5: Pushing the frontier with advanced reasoning, multimodality, long context, and next generation agentic capabilities.
\newblock Technical report, 2025.

\bibitem[Gong et~al.(2023)Gong, Lyu, Zhang, Wang, Zheng, Zhao, Liu, Zhang, Luo, and Chen]{gong2023multimodal}
Tao Gong, Chengqi Lyu, Shilong Zhang, Yudong Wang, Miao Zheng, Qian Zhao, Kuikun Liu, Wenwei Zhang, Ping Luo, and Kai Chen.
\newblock Multimodal-gpt: A vision and language model for dialogue with humans.
\newblock \emph{arXiv preprint arXiv:2305.04790}, 2023.

\bibitem[Grauman et~al.(2022)Grauman, Westbury, Byrne, Chavis, Furnari, Girdhar, Hamburger, Jiang, Liu, Liu, et~al.]{grauman2022ego4d}
Kristen Grauman, Andrew Westbury, Eugene Byrne, Zachary Chavis, Antonino Furnari, Rohit Girdhar, Jackson Hamburger, Hao Jiang, Miao Liu, Xingyu Liu, et~al.
\newblock Ego4d: Around the world in 3,000 hours of egocentric video.
\newblock In \emph{Proceedings of the IEEE/CVF conference on computer vision and pattern recognition}, pages 18995--19012, 2022.

\bibitem[He et~al.(2024{\natexlab{a}})He, Feng, Zheng, Lu, Zhu, Li, Fan, Wang, Li, Yang, et~al.]{he2024mmworld}
Xuehai He, Weixi Feng, Kaizhi Zheng, Yujie Lu, Wanrong Zhu, Jiachen Li, Yue Fan, Jianfeng Wang, Linjie Li, Zhengyuan Yang, et~al.
\newblock Mmworld: Towards multi-discipline multi-faceted world model evaluation in videos.
\newblock \emph{arXiv preprint arXiv:2406.08407}, 2024{\natexlab{a}}.

\bibitem[He et~al.(2024{\natexlab{b}})He, Wang, Yang, Wu, Wang, Wang, Zhan, Ruwase, Shen, and Wang]{he2024mojito}
Xuehai He, Shuohang Wang, Jianwei Yang, Xiaoxia Wu, Yiping Wang, Kuan Wang, Zheng Zhan, Olatunji Ruwase, Yelong Shen, and Xin~Eric Wang.
\newblock Mojito: Motion trajectory and intensity control for video generation.
\newblock \emph{arXiv preprint arXiv: 2412.08948}, 2024{\natexlab{b}}.

\bibitem[Hurst et~al.(2024)Hurst, Lerer, Goucher, Perelman, Ramesh, Clark, Ostrow, Welihinda, Hayes, Radford, et~al.]{hurst2024gpt}
Aaron Hurst, Adam Lerer, Adam~P Goucher, Adam Perelman, Aditya Ramesh, Aidan Clark, AJ Ostrow, Akila Welihinda, Alan Hayes, Alec Radford, et~al.
\newblock Gpt-4o system card.
\newblock \emph{arXiv preprint arXiv:2410.21276}, 2024.

\bibitem[Johansson(1973)]{Johansson1973Visual}
Gunnar Johansson.
\newblock Visual perception of biological motion and a model for its analysis.
\newblock \emph{Perception \& Psychophysics}, 14\penalty0 (2):\penalty0 201--211, 1973.

\bibitem[Khattak et~al.(2024)Khattak, Naeem, Hassan, Naseer, Tombari, Khan, and Khan]{khattak2024good}
Muhammad~Uzair Khattak, Muhammad~Ferjad Naeem, Jameel Hassan, Muzammal Naseer, Federico Tombari, Fahad~Shahbaz Khan, and Salman Khan.
\newblock How good is my video lmm? complex video reasoning and robustness evaluation suite for video-lmms.
\newblock \emph{arXiv preprint arXiv:2405.03690}, 2024.

\bibitem[Kim et~al.(2024)Kim, Pertsch, Karamcheti, Xiao, Balakrishna, Nair, Rafailov, Foster, Lam, Sanketi, et~al.]{kim2024openvla}
Moo~Jin Kim, Karl Pertsch, Siddharth Karamcheti, Ted Xiao, Ashwin Balakrishna, Suraj Nair, Rafael Rafailov, Ethan Foster, Grace Lam, Pannag Sanketi, et~al.
\newblock Openvla: An open-source vision-language-action model.
\newblock \emph{arXiv preprint arXiv:2406.09246}, 2024.

\bibitem[Kobayashi et~al.(2022)Kobayashi, Matsumoto, and Sitzmann]{kobayashi2022decomposing}
Sosuke Kobayashi, Eiichi Matsumoto, and Vincent Sitzmann.
\newblock Decomposing nerf for editing via feature field distillation.
\newblock \emph{Advances in neural information processing systems}, 35:\penalty0 23311--23330, 2022.

\bibitem[Laptev(2005)]{Laptev2005}
Ivan Laptev.
\newblock On space-time interest points.
\newblock \emph{International Journal of Computer Vision}, 64\penalty0 (2-3):\penalty0 107--123, 2005.

\bibitem[LeCun et~al.(1989)LeCun, Boser, Denker, Henderson, Howard, Hubbard, and Jackel]{LeCun1989Backpropagation}
Yann LeCun, Bernhard Boser, John~S. Denker, Donnie Henderson, Richard~E. Howard, Wayne Hubbard, and Lawrence~D. Jackel.
\newblock Backpropagation applied to handwritten zip code recognition.
\newblock \emph{Neural Computation}, 1\penalty0 (4):\penalty0 541--551, 1989.

\bibitem[Leslie(1984)]{Leslie1984Spatiotemporal}
Alan~M. Leslie.
\newblock Spatiotemporal continuity and the perception of causality in infants.
\newblock \emph{Perception}, 13\penalty0 (3):\penalty0 287--305, 1984.

\bibitem[Li et~al.(2024{\natexlab{a}})Li, Ge, Ge, Wang, Wang, Zhang, and Shan]{li2024seed}
Bohao Li, Yuying Ge, Yixiao Ge, Guangzhi Wang, Rui Wang, Ruimao Zhang, and Ying Shan.
\newblock Seed-bench: Benchmarking multimodal large language models.
\newblock In \emph{Proceedings of the IEEE/CVF Conference on Computer Vision and Pattern Recognition}, pages 13299--13308, 2024{\natexlab{a}}.

\bibitem[Li et~al.(2024{\natexlab{b}})Li, Zhang, Guo, Zhang, Li, Zhang, Zhang, Zhang, Li, Liu, et~al.]{li2024llava}
Bo Li, Yuanhan Zhang, Dong Guo, Renrui Zhang, Feng Li, Hao Zhang, Kaichen Zhang, Peiyuan Zhang, Yanwei Li, Ziwei Liu, et~al.
\newblock Llava-onevision: Easy visual task transfer.
\newblock \emph{arXiv preprint arXiv:2408.03326}, 2024{\natexlab{b}}.

\bibitem[Li et~al.(2024{\natexlab{c}})Li, Liu, Wu, Wang, Shen, Qu, Niu, Wang, Chen, and Li]{li2024aria}
Dongxu Li, Yudong Liu, Haoning Wu, Yue Wang, Zhiqi Shen, Bowen Qu, Xinyao Niu, Guoyin Wang, Bei Chen, and Junnan Li.
\newblock Aria: An open multimodal native mixture-of-experts model.
\newblock \emph{arXiv preprint arXiv:2410.05993}, 2024{\natexlab{c}}.

\bibitem[Li et~al.(2023{\natexlab{a}})Li, He, Wang, Li, Wang, Luo, Wang, Wang, and Qiao]{li2023videochat}
KunChang Li, Yinan He, Yi Wang, Yizhuo Li, Wenhai Wang, Ping Luo, Yali Wang, Limin Wang, and Yu Qiao.
\newblock Videochat: Chat-centric video understanding.
\newblock \emph{arXiv preprint arXiv:2305.06355}, 2023{\natexlab{a}}.

\bibitem[Li et~al.(2023{\natexlab{b}})Li, Wang, He, Li, Wang, Liu, Wang, Xu, Chen, Luo, Wang, and Qiao]{li2023mvbench}
Kunchang Li, Yali Wang, Yinan He, Yizhuo Li, Yi Wang, Yi Liu, Zun Wang, Jilan Xu, Guo Chen, Ping Luo, Limin Wang, and Yu Qiao.
\newblock Mvbench: A comprehensive multi-modal video understanding benchmark, 2023{\natexlab{b}}.

\bibitem[Li et~al.(2024{\natexlab{d}})Li, Wang, He, Li, Wang, Liu, Wang, Xu, Chen, Luo, et~al.]{li2024mvbench}
Kunchang Li, Yali Wang, Yinan He, Yizhuo Li, Yi Wang, Yi Liu, Zun Wang, Jilan Xu, Guo Chen, Ping Luo, et~al.
\newblock Mvbench: A comprehensive multi-modal video understanding benchmark.
\newblock In \emph{Proceedings of the IEEE/CVF Conference on Computer Vision and Pattern Recognition}, pages 22195--22206, 2024{\natexlab{d}}.

\bibitem[Li et~al.(2025)Li, Pan, Yang, Xu, Zhou, Zhang, Li, Kadambi, Wang, Tu, and Fan]{li2025kdgen}
Renjie Li, Panwang Pan, Bangbang Yang, Dejia Xu, Shijie Zhou, Xuanyang Zhang, Zeming Li, Achuta Kadambi, Zhangyang Wang, Zhengzhong Tu, and Zhiwen Fan.
\newblock 4k4{DG}en: Panoramic 4d generation at 4k resolution.
\newblock In \emph{The Thirteenth International Conference on Learning Representations}, 2025.

\bibitem[Li et~al.(2024{\natexlab{e}})Li, Huang, Wang, Li, and Wang]{li2024videoeval}
Xinhao Li, Zhenpeng Huang, Jing Wang, Kunchang Li, and Limin Wang.
\newblock Videoeval: Comprehensive benchmark suite for low-cost evaluation of video foundation model.
\newblock \emph{arXiv preprint arXiv:2407.06491}, 2024{\natexlab{e}}.

\bibitem[Ling et~al.(2025)Ling, Lin, Lin, Ding, Zeng, Sheng, Ge, Liu, Bera, and Li]{ling2025scenethesis}
Lu Ling, Chen-Hsuan Lin, Tsung-Yi Lin, Yifan Ding, Yu Zeng, Yichen Sheng, Yunhao Ge, Ming-Yu Liu, Aniket Bera, and Zhaoshuo Li.
\newblock Scenethesis: A language and vision agentic framework for 3d scene generation.
\newblock \emph{arXiv preprint arXiv:2505.02836}, 2025.

\bibitem[Liu et~al.(2023)Liu, Li, Wu, and Lee]{liu2023visual}
Haotian Liu, Chunyuan Li, Qingyang Wu, and Yong~Jae Lee.
\newblock Visual instruction tuning.
\newblock \emph{Advances in neural information processing systems}, 36:\penalty0 34892--34916, 2023.

\bibitem[Liu et~al.(2024{\natexlab{a}})Liu, Yan, Zaharia, and Abbeel]{liu2023world}
Hao Liu, Wilson Yan, Matei Zaharia, and Pieter Abbeel.
\newblock World model on million-length video and language with ringattention.
\newblock \emph{arXiv preprint}, 2024{\natexlab{a}}.

\bibitem[Liu et~al.(2024{\natexlab{b}})Liu, Yan, Zaharia, and Abbeel]{liu2024world}
Hao Liu, Wilson Yan, Matei Zaharia, and Pieter Abbeel.
\newblock World model on million-length video and language with blockwise ringattention.
\newblock \emph{arXiv preprint arXiv:2402.08268}, 2024{\natexlab{b}}.

\bibitem[Liu et~al.(2024{\natexlab{c}})Liu, Duan, Zhang, Li, Zhang, Zhao, Yuan, Wang, He, Liu, et~al.]{liu2024mmbench}
Yuan Liu, Haodong Duan, Yuanhan Zhang, Bo Li, Songyang Zhang, Wangbo Zhao, Yike Yuan, Jiaqi Wang, Conghui He, Ziwei Liu, et~al.
\newblock Mmbench: Is your multi-modal model an all-around player?
\newblock In \emph{European conference on computer vision}, pages 216--233. Springer, 2024{\natexlab{c}}.

\bibitem[Lu et~al.(2024)Lu, Liu, Zhang, Wang, Dong, Liu, Sun, Ren, Li, Yang, et~al.]{lu2024deepseek}
Haoyu Lu, Wen Liu, Bo Zhang, Bingxuan Wang, Kai Dong, Bo Liu, Jingxiang Sun, Tongzheng Ren, Zhuoshu Li, Hao Yang, et~al.
\newblock Deepseek-vl: towards real-world vision-language understanding.
\newblock \emph{arXiv preprint arXiv:2403.05525}, 2024.

\bibitem[Lu et~al.(2019)Lu, Batra, Parikh, and Lee]{Lu2019ViLBERTPT}
Jiasen Lu, Dhruv Batra, Devi Parikh, and Stefan Lee.
\newblock Vilbert: Pretraining task-agnostic visiolinguistic representations for vision-and-language tasks.
\newblock In \emph{Neural Information Processing Systems}, 2019.

\bibitem[Maaz et~al.(2023)Maaz, Rasheed, Khan, and Khan]{maaz2023video}
Muhammad Maaz, Hanoona Rasheed, Salman Khan, and Fahad~Shahbaz Khan.
\newblock Video-chatgpt: Towards detailed video understanding via large vision and language models.
\newblock \emph{arXiv preprint arXiv:2306.05424}, 2023.

\bibitem[Marr and Ullman(1981)]{Marr1981Directional}
D. Marr and S. Ullman.
\newblock Directional selectivity and its use in early visual processing.
\newblock \emph{Proceedings of the Royal Society of London. Series B, Biological Sciences}, 211\penalty0 (1183):\penalty0 151--180, 1981.

\bibitem[Meta(2025)]{llama4}
Meta.
\newblock The llama 4 herd: The beginning of a new era of natively multimodal ai innovation.
\newblock Technical report, 2025.

\bibitem[Ning et~al.(2023)Ning, Zhu, Xie, Lin, Cui, Yuan, Chen, and Yuan]{ning2023video}
Munan Ning, Bin Zhu, Yujia Xie, Bin Lin, Jiaxi Cui, Lu Yuan, Dongdong Chen, and Li Yuan.
\newblock Video-bench: A comprehensive benchmark and toolkit for evaluating video-based large language models.
\newblock \emph{arXiv preprint arXiv:2311.16103}, 2023.

\bibitem[{\"O}cal et~al.(2024){\"O}cal, Tatarchenko, Karao{\u{g}}lu, and Gevers]{ocal2024sceneteller}
Ba{\c{s}}ak~Melis {\"O}cal, Maxim Tatarchenko, Sezer Karao{\u{g}}lu, and Theo Gevers.
\newblock Sceneteller: Language-to-3d scene generation.
\newblock In \emph{European Conference on Computer Vision}, pages 362--378. Springer, 2024.

\bibitem[OpenAI(2024)]{gpt4o}
OpenAI.
\newblock Hello gpt-4o.
\newblock Technical report, 2024.

\bibitem[Patel et~al.(2025)Patel, Yin, Huang, Garg, Nayyeri, Fei-Fei, Lazebnik, and Li]{patel2025real}
Shivansh Patel, Xinchen Yin, Wenlong Huang, Shubham Garg, Hooshang Nayyeri, Li Fei-Fei, Svetlana Lazebnik, and Yunzhu Li.
\newblock A real-to-sim-to-real approach to robotic manipulation with vlm-generated iterative keypoint rewards.
\newblock \emph{arXiv preprint arXiv:2502.08643}, 2025.

\bibitem[Perazzi et~al.(2016)Perazzi, Pont-Tuset, McWilliams, Van~Gool, Gross, and Sorkine-Hornung]{perazzi2016benchmark}
Federico Perazzi, Jordi Pont-Tuset, Brian McWilliams, Luc Van~Gool, Markus Gross, and Alexander Sorkine-Hornung.
\newblock A benchmark dataset and evaluation methodology for video object segmentation.
\newblock In \emph{Proceedings of the IEEE conference on computer vision and pattern recognition}, pages 724--732, 2016.

\bibitem[Pont-Tuset et~al.(2017)Pont-Tuset, Perazzi, Caelles, Arbel{\'a}ez, Sorkine-Hornung, and Van~Gool]{davis2017}
Jordi Pont-Tuset, Federico Perazzi, Sergi Caelles, Pablo Arbel{\'a}ez, Alex Sorkine-Hornung, and Luc Van~Gool.
\newblock The 2017 davis challenge on video object segmentation.
\newblock \emph{arXiv preprint arXiv:1704.00675}, 2017.

\bibitem[Radford et~al.(2018)Radford, Narasimhan, Salimans, Sutskever, et~al.]{radford2018improving}
Alec Radford, Karthik Narasimhan, Tim Salimans, Ilya Sutskever, et~al.
\newblock Improving language understanding by generative pre-training.
\newblock 2018.

\bibitem[Radford et~al.(2021)Radford, Kim, Hallacy, Ramesh, Goh, Agarwal, Sastry, Askell, Mishkin, Clark, Krueger, and Sutskever]{Radford2021Learning}
Alec Radford, Jong~Wook Kim, Chris Hallacy, Aditya Ramesh, Gabriel Goh, Sandhini Agarwal, Girish Sastry, Amanda Askell, Pamela Mishkin, Jack Clark, Gretchen Krueger, and Ilya Sutskever.
\newblock Learning transferable visual models from natural language supervision.
\newblock In \emph{Proceedings of the 38th International Conference on Machine Learning}, pages 8748--8763, 2021.

\bibitem[Ranasinghe et~al.(2024)Ranasinghe, Shukla, Poursaeed, Ryoo, and Lin]{ranasinghe2024learning}
Kanchana Ranasinghe, Satya~Narayan Shukla, Omid Poursaeed, Michael~S Ryoo, and Tsung-Yu Lin.
\newblock Learning to localize objects improves spatial reasoning in visual-llms.
\newblock In \emph{Proceedings of the IEEE/CVF Conference on Computer Vision and Pattern Recognition}, pages 12977--12987, 2024.

\bibitem[Simonyan and Zisserman(2014)]{Simonyan2014}
Karen Simonyan and Andrew Zisserman.
\newblock Two-stream convolutional networks for action recognition in videos.
\newblock In \emph{Advances in Neural Information Processing Systems (NIPS)}, pages 568--576, 2014.

\bibitem[Spelke and Kinzler(2007)]{Spelke2007Core}
Elizabeth~S. Spelke and Katherine~D. Kinzler.
\newblock Core knowledge.
\newblock \emph{Developmental Science}, 10\penalty0 (1):\penalty0 89--96, 2007.

\bibitem[Suglia et~al.(2024)Suglia, Greco, Baker, Part, Papaioannou, Eshghi, Konstas, and Lemon]{suglia2024alanavlm}
Alessandro Suglia, Claudio Greco, Katie Baker, Jose~L Part, Ioannis Papaioannou, Arash Eshghi, Ioannis Konstas, and Oliver Lemon.
\newblock Alanavlm: A multimodal embodied ai foundation model for egocentric video understanding.
\newblock \emph{arXiv preprint arXiv:2406.13807}, 2024.

\bibitem[Team et~al.(2024)Team, Georgiev, Lei, Burnell, Bai, Gulati, Tanzer, Vincent, Pan, Wang, et~al.]{team2024gemini}
Gemini Team, Petko Georgiev, Ving~Ian Lei, Ryan Burnell, Libin Bai, Anmol Gulati, Garrett Tanzer, Damien Vincent, Zhufeng Pan, Shibo Wang, et~al.
\newblock Gemini 1.5: Unlocking multimodal understanding across millions of tokens of context.
\newblock \emph{arXiv preprint arXiv:2403.05530}, 2024.

\bibitem[Touvron et~al.(2023)Touvron, Lavril, Izacard, Martinet, Lachaux, Lacroix, Rozi{\`e}re, Goyal, Hambro, Azhar, et~al.]{touvron2023llama}
Hugo Touvron, Thibaut Lavril, Gautier Izacard, Xavier Martinet, Marie-Anne Lachaux, Timoth{\'e}e Lacroix, Baptiste Rozi{\`e}re, Naman Goyal, Eric Hambro, Faisal Azhar, et~al.
\newblock Llama: Open and efficient foundation language models.
\newblock \emph{arXiv preprint arXiv:2302.13971}, 2023.

\bibitem[Wan et~al.(2025)Wan, Wang, Ai, Wen, Mao, Xie, Chen, Yu, Zhao, Yang, et~al.]{wan2025wan}
Team Wan, Ang Wang, Baole Ai, Bin Wen, Chaojie Mao, Chen-Wei Xie, Di Chen, Feiwu Yu, Haiming Zhao, Jianxiao Yang, et~al.
\newblock Wan: Open and advanced large-scale video generative models.
\newblock \emph{arXiv preprint arXiv:2503.20314}, 2025.

\bibitem[Wang et~al.(2024{\natexlab{a}})Wang, Zhang, Dong, Fang, and Feng]{wang2024vlm}
Beichen Wang, Juexiao Zhang, Shuwen Dong, Irving Fang, and Chen Feng.
\newblock Vlm see, robot do: Human demo video to robot action plan via vision language model.
\newblock \emph{arXiv preprint arXiv:2410.08792}, 2024{\natexlab{a}}.

\bibitem[Wang et~al.(2011)Wang, Kl{\"a}ser, Schmid, and Liu]{Wang2011}
Heng Wang, Alexander Kl{\"a}ser, Cordelia Schmid, and Cheng-Lin Liu.
\newblock Action recognition by dense trajectories.
\newblock In \emph{2011 IEEE Conference on Computer Vision and Pattern Recognition (CVPR)}, pages 3169--3176, 2011.

\bibitem[Wang et~al.(2024{\natexlab{b}})Wang, Xu, Cheng, Diao, Zhou, Cao, Wang, Ge, and Huang]{wang2024grounded}
Haibo Wang, Zhiyang Xu, Yu Cheng, Shizhe Diao, Yufan Zhou, Yixin Cao, Qifan Wang, Weifeng Ge, and Lifu Huang.
\newblock Grounded-videollm: Sharpening fine-grained temporal grounding in video large language models.
\newblock \emph{arXiv preprint arXiv:2410.03290}, 2024{\natexlab{b}}.

\bibitem[Wang et~al.(2024{\natexlab{c}})Wang, Bai, Tan, Wang, Fan, Bai, Chen, Liu, Wang, Ge, et~al.]{wang2024qwen2}
Peng Wang, Shuai Bai, Sinan Tan, Shijie Wang, Zhihao Fan, Jinze Bai, Keqin Chen, Xuejing Liu, Jialin Wang, Wenbin Ge, et~al.
\newblock Qwen2-vl: Enhancing vision-language model's perception of the world at any resolution.
\newblock \emph{arXiv preprint arXiv:2409.12191}, 2024{\natexlab{c}}.

\bibitem[Wang et~al.(2024{\natexlab{d}})Wang, Ma, Wang, Chen, Kortylewski, and Yuille]{wang2024compositional}
Xingrui Wang, Wufei Ma, Angtian Wang, Shuo Chen, Adam Kortylewski, and Alan Yuille.
\newblock Compositional 4d dynamic scenes understanding with physics priors for video question answering.
\newblock \emph{arXiv preprint arXiv:2406.00622}, 2024{\natexlab{d}}.

\bibitem[Wang et~al.(2024{\natexlab{e}})Wang, Li, Li, Yu, He, Chen, Pei, Zheng, Wang, Shi, et~al.]{wang2024internvideo2}
Yi Wang, Kunchang Li, Xinhao Li, Jiashuo Yu, Yinan He, Guo Chen, Baoqi Pei, Rongkun Zheng, Zun Wang, Yansong Shi, et~al.
\newblock Internvideo2: Scaling foundation models for multimodal video understanding.
\newblock In \emph{European Conference on Computer Vision}, pages 396--416. Springer, 2024{\natexlab{e}}.

\bibitem[Wang et~al.(2025)Wang, Li, Yan, He, Yu, Zeng, Wang, Ma, Huang, Gao, et~al.]{wang2025internvideo2}
Yi Wang, Xinhao Li, Ziang Yan, Yinan He, Jiashuo Yu, Xiangyu Zeng, Chenting Wang, Changlian Ma, Haian Huang, Jianfei Gao, et~al.
\newblock Internvideo2. 5: Empowering video mllms with long and rich context modeling.
\newblock \emph{arXiv preprint arXiv:2501.12386}, 2025.

\bibitem[Wei et~al.(2021)Wei, Bosma, Zhao, Guu, Yu, Lester, Du, Dai, and Le]{wei2021finetuned}
Jason Wei, Maarten Bosma, Vincent~Y Zhao, Kelvin Guu, Adams~Wei Yu, Brian Lester, Nan Du, Andrew~M Dai, and Quoc~V Le.
\newblock Finetuned language models are zero-shot learners.
\newblock \emph{arXiv preprint arXiv:2109.01652}, 2021.

\bibitem[Wei et~al.(2022)Wei, Wang, Schuurmans, Bosma, Xia, Chi, Le, Zhou, et~al.]{wei2022chain}
Jason Wei, Xuezhi Wang, Dale Schuurmans, Maarten Bosma, Fei Xia, Ed Chi, Quoc~V Le, Denny Zhou, et~al.
\newblock Chain-of-thought prompting elicits reasoning in large language models.
\newblock \emph{Advances in neural information processing systems}, 35:\penalty0 24824--24837, 2022.

\bibitem[Wu et~al.(2025)Wu, Gao, Poole, Trevithick, Zheng, Barron, and Holynski]{wu2025cat4d}
Rundi Wu, Ruiqi Gao, Ben Poole, Alex Trevithick, Changxi Zheng, Jonathan~T Barron, and Aleksander Holynski.
\newblock Cat4d: Create anything in 4d with multi-view video diffusion models.
\newblock In \emph{Proceedings of the Computer Vision and Pattern Recognition Conference}, pages 26057--26068, 2025.

\bibitem[Wu et~al.(2024)Wu, Chen, Lin, Wang, Gao, Xu, Xu, Hu, Chen, and Shou]{wu2024videollm}
Shiwei Wu, Joya Chen, Kevin~Qinghong Lin, Qimeng Wang, Yan Gao, Qianli Xu, Tong Xu, Yao Hu, Enhong Chen, and Mike~Zheng Shou.
\newblock Videollm-mod: Efficient video-language streaming with mixture-of-depths vision computation.
\newblock \emph{Advances in Neural Information Processing Systems}, 37:\penalty0 109922--109947, 2024.

\bibitem[xAI(2024)]{grok2}
xAI.
\newblock Grok-2 beta release.
\newblock Technical report, 2024.

\bibitem[Xu et~al.(2018)Xu, Yang, Fan, Yang, Yue, Liang, Price, Cohen, and Huang]{xu2018youtube}
Ning Xu, Linjie Yang, Yuchen Fan, Jianchao Yang, Dingcheng Yue, Yuchen Liang, Brian Price, Scott Cohen, and Thomas Huang.
\newblock Youtube-vos: Sequence-to-sequence video object segmentation.
\newblock In \emph{Proceedings of the European conference on computer vision (ECCV)}, pages 585--601, 2018.

\bibitem[Yang et~al.(2024{\natexlab{a}})Yang, Yang, Zhang, Hui, Zheng, Yu, Li, Liu, Huang, Wei, et~al.]{yang2024qwen2}
An Yang, Baosong Yang, Beichen Zhang, Binyuan Hui, Bo Zheng, Bowen Yu, Chengyuan Li, Dayiheng Liu, Fei Huang, Haoran Wei, et~al.
\newblock Qwen2. 5 technical report.
\newblock \emph{arXiv preprint arXiv:2412.15115}, 2024{\natexlab{a}}.

\bibitem[Yang et~al.(2024{\natexlab{b}})Yang, Yang, Gupta, Han, Fei-Fei, and Xie]{yang2024thinking}
Jihan Yang, Shusheng Yang, Anjali~W Gupta, Rilyn Han, Li Fei-Fei, and Saining Xie.
\newblock Thinking in space: How multimodal large language models see, remember, and recall spaces.
\newblock \emph{arXiv preprint arXiv:2412.14171}, 2024{\natexlab{b}}.

\bibitem[Yuan et~al.(2024)Yuan, Zhang, Li, Cheng, Zhang, Li, Li, Zhao, Zhang, Zhuang, et~al.]{yuan2024videorefer}
Yuqian Yuan, Hang Zhang, Wentong Li, Zesen Cheng, Boqiang Zhang, Long Li, Xin Li, Deli Zhao, Wenqiao Zhang, Yueting Zhuang, et~al.
\newblock Videorefer suite: Advancing spatial-temporal object understanding with video llm.
\newblock \emph{arXiv preprint arXiv:2501.00599}, 2024.

\bibitem[Yue et~al.(2024{\natexlab{a}})Yue, Ni, Zhang, Zheng, Liu, Zhang, Stevens, Jiang, Ren, Sun, et~al.]{yue2024mmmu}
Xiang Yue, Yuansheng Ni, Kai Zhang, Tianyu Zheng, Ruoqi Liu, Ge Zhang, Samuel Stevens, Dongfu Jiang, Weiming Ren, Yuxuan Sun, et~al.
\newblock Mmmu: A massive multi-discipline multimodal understanding and reasoning benchmark for expert agi.
\newblock In \emph{Proceedings of the IEEE/CVF Conference on Computer Vision and Pattern Recognition}, pages 9556--9567, 2024{\natexlab{a}}.

\bibitem[Yue et~al.(2024{\natexlab{b}})Yue, Das, Engelmann, Tang, and Lenssen]{yue2024improving}
Yuanwen Yue, Anurag Das, Francis Engelmann, Siyu Tang, and Jan~Eric Lenssen.
\newblock Improving 2d feature representations by 3d-aware fine-tuning.
\newblock In \emph{European Conference on Computer Vision}, pages 57--74. Springer, 2024{\natexlab{b}}.

\bibitem[Zhang et~al.(2025)Zhang, Li, Cheng, Hu, Yuan, Chen, Leng, Jiang, Zhang, Li, et~al.]{zhang2025videollama}
Boqiang Zhang, Kehan Li, Zesen Cheng, Zhiqiang Hu, Yuqian Yuan, Guanzheng Chen, Sicong Leng, Yuming Jiang, Hang Zhang, Xin Li, et~al.
\newblock Videollama 3: Frontier multimodal foundation models for image and video understanding.
\newblock \emph{arXiv preprint arXiv:2501.13106}, 2025.

\bibitem[Zhang et~al.(2023)Zhang, Li, and Bing]{zhang2023video}
Hang Zhang, Xin Li, and Lidong Bing.
\newblock Video-llama: An instruction-tuned audio-visual language model for video understanding.
\newblock \emph{arXiv preprint arXiv:2306.02858}, 2023.

\bibitem[Zhang et~al.(2024{\natexlab{a}})Zhang, Wang, Lyu, Zhang, Chen, Shu, Du, and Gan]{zhang2024combo}
Hongxin Zhang, Zeyuan Wang, Qiushi Lyu, Zheyuan Zhang, Sunli Chen, Tianmin Shu, Yilun Du, and Chuang Gan.
\newblock Combo: compositional world models for embodied multi-agent cooperation.
\newblock \emph{arXiv preprint arXiv:2404.10775}, 2024{\natexlab{a}}.

\bibitem[Zhang et~al.(2024{\natexlab{b}})Zhang, Li, Liu, Lee, Gui, Fu, Feng, Liu, and Li]{zhang2024llavanextvideo}
Yuanhan Zhang, Bo Li, haotian Liu, Yong~jae Lee, Liangke Gui, Di Fu, Jiashi Feng, Ziwei Liu, and Chunyuan Li.
\newblock Llava-next: A strong zero-shot video understanding model, 2024{\natexlab{b}}.

\bibitem[Zhao et~al.(2025)Zhao, Xie, Zhang, Gan, Long, Hu, Hu, Chen, Li, Song, et~al.]{zhao2025mmvu}
Yilun Zhao, Lujing Xie, Haowei Zhang, Guo Gan, Yitao Long, Zhiyuan Hu, Tongyan Hu, Weiyuan Chen, Chuhan Li, Junyang Song, et~al.
\newblock Mmvu: Measuring expert-level multi-discipline video understanding.
\newblock \emph{arXiv preprint arXiv:2501.12380}, 2025.

\bibitem[Zheng et~al.(2024)Zheng, Zhang, Zhang, Ye, Luo, Feng, and Ma]{zheng2024llamafactory}
Yaowei Zheng, Richong Zhang, Junhao Zhang, Yanhan Ye, Zheyan Luo, Zhangchi Feng, and Yongqiang Ma.
\newblock Llamafactory: Unified efficient fine-tuning of 100+ language models.
\newblock In \emph{Proceedings of the 62nd Annual Meeting of the Association for Computational Linguistics (Volume 3: System Demonstrations)}, Bangkok, Thailand, 2024. Association for Computational Linguistics.

\bibitem[Zhou et~al.(2024{\natexlab{a}})Zhou, Chang, Jiang, Fan, Zhu, Xu, Chari, You, Wang, and Kadambi]{zhou2024feature}
Shijie Zhou, Haoran Chang, Sicheng Jiang, Zhiwen Fan, Zehao Zhu, Dejia Xu, Pradyumna Chari, Suya You, Zhangyang Wang, and Achuta Kadambi.
\newblock Feature 3dgs: Supercharging 3d gaussian splatting to enable distilled feature fields.
\newblock In \emph{Proceedings of the IEEE/CVF Conference on Computer Vision and Pattern Recognition}, pages 21676--21685, 2024{\natexlab{a}}.

\bibitem[Zhou et~al.(2024{\natexlab{b}})Zhou, Fan, Xu, Chang, Chari, Bharadwaj, You, Wang, and Kadambi]{zhou2024dreamscene360}
Shijie Zhou, Zhiwen Fan, Dejia Xu, Haoran Chang, Pradyumna Chari, Tejas Bharadwaj, Suya You, Zhangyang Wang, and Achuta Kadambi.
\newblock Dreamscene360: Unconstrained text-to-3d scene generation with panoramic gaussian splatting.
\newblock In \emph{European Conference on Computer Vision}, pages 324--342. Springer, 2024{\natexlab{b}}.

\bibitem[Zhou et~al.(2025)Zhou, Ren, Weng, Zhang, Wang, Xu, Fan, You, Wang, Guibas, et~al.]{zhou2025feature4x}
Shijie Zhou, Hui Ren, Yijia Weng, Shuwang Zhang, Zhen Wang, Dejia Xu, Zhiwen Fan, Suya You, Zhangyang Wang, Leonidas Guibas, et~al.
\newblock Feature4x: Bridging any monocular video to 4d agentic ai with versatile gaussian feature fields.
\newblock In \emph{Proceedings of the Computer Vision and Pattern Recognition Conference}, pages 14179--14190, 2025.

\bibitem[Zhu et~al.(2023)Zhu, Chen, Shen, Li, and Elhoseiny]{zhu2023minigpt}
Deyao Zhu, Jun Chen, Xiaoqian Shen, Xiang Li, and Mohamed Elhoseiny.
\newblock Minigpt-4: Enhancing vision-language understanding with advanced large language models.
\newblock \emph{arXiv preprint arXiv:2304.10592}, 2023.

\bibitem[Zohar et~al.(2024)Zohar, Wang, Dubois, Mehta, Xiao, Hansen-Estruch, Yu, Wang, Juefei-Xu, Zhang, et~al.]{zohar2024apollo}
Orr Zohar, Xiaohan Wang, Yann Dubois, Nikhil Mehta, Tong Xiao, Philippe Hansen-Estruch, Licheng Yu, Xiaofang Wang, Felix Juefei-Xu, Ning Zhang, et~al.
\newblock Apollo: An exploration of video understanding in large multimodal models.
\newblock \emph{arXiv preprint arXiv:2412.10360}, 2024.

\end{thebibliography}
